\documentclass{iopjournal}

\usepackage{amsmath,amssymb,amsthm,mathtools}
\usepackage{booktabs}
\usepackage{array}
\usepackage{graphicx}
\usepackage{enumitem}
\usepackage{hyperref}
\usepackage[nameinlink,capitalize]{cleveref}
\usepackage[style=numeric, sorting=none]{biblatex}
\usepackage{comment}
\usepackage{subcaption}
\usepackage{braket}
\usepackage{xcolor}
\usepackage{multirow}
\usepackage{float}

\hypersetup{
  colorlinks=true,    
  linkcolor=blue,
  citecolor=blue,
  urlcolor=blue
}

\newtheorem{theorem}{Theorem}[section]
\newtheorem{proposition}[theorem]{Proposition}

\theoremstyle{definition}

\theoremstyle{remark}

\newcommand{\dd}{\mathrm{d}}
\newcommand{\id}{\mathrm{id}}

\newcommand{\OO}{\mathrm{O}}
\newcommand{\GL}{\mathrm{GL}}

\begin{document}

\title{Enforcing Symmetries on Pre-trained ML Models for Physics}

\author{Wenjie Zhou$^1$\orcid{0009-0009-0332-0812}, Valentino F.\ Foit$^2$\orcid{0000-0002-7864-7053}, David W.\ Hogg$^{2,3,4}$\orcid{0000-0003-2866-9403}, and Soledad Villar$^{5,6, *}$\orcid{0000-0003-4968-3829}}

\affil{$^1$Department of Physics and Astronomy, Johns Hopkins University, Baltimore, MD 21218, USA}\\

\affil{$^2$Center for Cosmology and Particle Physics, Department of Physics, New York University, New York, NY 10003, USA}\\

\affil{$^3$Max-Planck-Institut f\"ur Astronomie, K\"onigstuhl 17, D-69117 Heidelberg, Germany}\\

\affil{$^4$Center for Computational Astrophysics, Flatiron Institute, 162 Fifth Ave., New York, NY 10010, USA}

\affil{$^5$Department of Applied Mathematics and Statistics and Mathematical Institute for Data Science, Johns Hopkins University, Baltimore, MD 21218, USA}

\affil{$^6$Center for Computational Mathematics, Flatiron Institute, 162 Fifth Ave., New York, NY 10010, USA}

\affil{$^*$Author to whom any correspondence should be addressed.}\\
\email{soledad.villar@jhu.edu}


\begin{abstract}

Post-hoc group averaging (or test-time augmentation) is a simple way to impose symmetry on a trained predictor without changing its architecture or retraining it. For finite and compact groups, averaging over a normalized Haar measure on the group gives the Reynolds operator: a canonical projection onto invariant or equivariant predictors with known non-increasing risk guarantees under certain assumptions. This paper provides numerical experiments on physics emulation models that demonstrate the effectiveness of the approach in this setting. We then extend this framework to non-compact group symmetries. For non-compact groups, the Haar measure cannot be normalized, so group averaging cannot be defined in the same way. We replace it with three alternatives: the Weyl unitary trick (which requires complexifying the space and extending the predictor), a data-dependent orthogonal projection (which involves a non-trivial optimization on the space of invariant or equivariant functions), and a frame averaging approach (which involves mapping each point to a set of orbit representatives). We illustrate the constructions theoretically and empirically on the unit-rescaling group and the Lorentz group. Our results show that in the non-compact case, the Weyl trick and frame averaging do not provide a consistent improvement over the baseline, except where the test data is out-of-distribution (an orders-of-magnitude improvement in this case) and, for the Weyl trick, in low-data regimes. The orthogonal projection can provide improvements in-distribution and out-of-distribution. We also observe that models trained directly on invariant features matched or beat the projection of an unconstrained one. We conclude that post-hoc symmetrization extends to non-compact groups, but it does not consistently outperform training a symmetric model.
\end{abstract}

\section{Introduction}

A common guiding principle in machine learning for the physical sciences is that predictive models should be consistent with the laws of physics. These laws are often manifestly symmetric, which constrains admissible observables and determines how physical quantities transform. When learning from data generated by a physical system, respecting these symmetries is not only conceptually natural, but can also improve robustness.

One way to incorporate symmetry into learning models is through invariant or equivariant architectures, which enforce symmetries by construction. Although such approaches have been successful in many settings, they are not always used in practice. Implementing equivariant models typically requires nontrivial representation-theoretic or invariant-theoretic machinery, and existing equivariant methods are not always compatible with state-of-the-art architectures. As a result, many modern machine learning models for physics prioritize well-understood, symmetry-agnostic architectures, relying on data to learn the physics rather than explicit symmetry constraints.

An alternative approach is to impose symmetry post hoc by projecting a trained model onto the space of symmetry-preserving functions \cite{lehmannCasella1998,ChenDobribanLee2020DataAugmentation,lyle2020benefits,sannai2024invariant}. In this framework, a pretrained predictor \( f \) is treated as an arbitrary function, and a new predictor \( \bar f \) is obtained by averaging \( f \) over group orbits. For compact groups, this procedure is canonically implemented by Reynolds averaging with respect to the normalized Haar measure of the group (a.k.a. the invariant volume measure). Theoretical work shows that this projection yields a provably nonnegative, and in some cases strictly positive, improvement in expected risk under invariance assumptions on the data distribution, without requiring retraining or architectural modification \cite{ElesedyZaidi2021}.   However, this result applies only to compact groups, without results for non-compact ones. Yet, many physical symmetries are non-compact. For instance, the Lorentz group plays a central role in particle physics.

In this work, we first present empirical evidence showing that the group averaging approach works in the compact case.  To this end, we take pretrained surrogate models released with The Well \cite{ohana2024well} and compare each with its group-averaged version. In many cases, this mere act of averaging, even when the group is small, already improves the results significantly. We then adapt the methodology to non-compact groups. We first argue why the approach and its guarantees fail, and we then investigate alternatives. 

The remainder of the paper is organized as follows. Section~\ref{sec:related_work} reviews related work on invariant and equivariant learning, data augmentation, and post-hoc symmetrization. Section~\ref{sec:compact-reynolds} introduces Reynolds averaging for compact groups, recalls its risk-improvement guarantee, and evaluates it on pretrained foundation models for physics. Section~\ref{sec:lorentz-weyl} explains the obstruction posed by non-compact groups and develops projection methods for unit rescalings and Lorentz transformations, together with numerical experiments. Section~\ref{sec:discussion-conclusion} discusses the limitations and concludes.

\section{Related work}
\label{sec:related_work}

Symmetry has long served as an inductive bias in machine learning for physical systems, where predictors are constrained by transformation laws; building symmetries into network structure dates back at least to \cite{shawe1989building}. A large body of work enforces these constraints directly in the architecture: group-equivariant convolutional networks \cite{cohen2016group}, their generalization to arbitrary compact group actions \cite{kondor2018generalization,cohen2019general}, steerable and irreducible-representation-based layers \cite{cohen2016steerable,kondor2018n,geiger2022e3nn}, equivariant multilayer perceptrons for arbitrary matrix groups, including the Lorentz group \(O(1,3)\) \cite{FinziWellingWilson2021EMLP}, and further exactly or approximately equivariant designs \cite{worrall2019deep,wang2020incorporating,huang2023approximately,gregory2024robust}. A related group-averaging construction appears in variational quantum machine learning, where twirling projects gate generators onto the commutant of a unitary symmetry representation to construct equivariant quantum circuits \cite{PRXQuantum.4.010328}. A complementary invariant-theoretic viewpoint characterizes equivariant maps by generating scalar invariants and tensorial output transformation laws \cite{villar2021scalars,blum2023machine,blum2024galois,gregory2024learning}. This perspective is especially natural in physics, where scalar contractions often determine the allowed form of a model. All of these methods fix the symmetry constraint before training, by modifying the model class or the input representation.

A different route is to impose symmetry after training. Averaging a predictor over the group orbit is the classical symmetrization device, used, for example, to establish universality of invariant models \cite{yarotsky2022universal} and to define permutation-invariant pooling \cite{murphy2019relational}. For compact groups, averaging with respect to the normalized Haar measure gives the Reynolds operator, which projects an arbitrary predictor onto the invariant or equivariant class. When the data distribution and target are \(G\)-invariant, this projection cannot increase squared risk and produces a strict improvement whenever the predictor has a nonzero non-equivariant component \cite{ElesedyZaidi2021}. Related theoretical work quantifies the generalization and sample-complexity benefits of enforcing symmetry \cite{petrache2023approximation,tahmasebi2023exact,bietti2021samplecomplexitylearninginvariance, tahmasebi2026achievingapproximatesymmetryexponentially}. 

On the empirical side, inexpensive post-hoc group averaging has been shown to improve already-trained surrogate models at test time, without retraining or architectural modification \cite{foit2025groupaveragingphysicsapplications}. Several related methods impose symmetry without averaging over the entire group. Frame averaging evaluates a non-equivariant backbone over a small, input-dependent set of canonical frames and averages the resulting predictions \cite{puny2022frameaveraginginvariantequivariant, lin2026equivarianceminimalframeaveraging}. Learned canonicalization instead learns a transformation that maps each input to a canonical representative, applies a non-equivariant backbone, and transforms the output back \cite{kaba2023equivariance}. Lie-algebra-based canonicalization extends this paradigm to continuous non-compact symmetry groups and can be applied to pre-trained neural operators \cite{shumaylov2025liealgebracanonicalizationequivariant}. Reynolds networks build the projection directly into the network architecture \cite{sannai2024invariant}. Probabilistic symmetrization instead learns an input-dependent probability distribution over group transformations. It evaluates the backbone in transformations sampled from this distribution, transforms the predictions back, and averages them. The distribution is designed to shift consistently when the input is transformed, making the exact expected prediction equivariant \cite{KimEtAl2023LPS}. Finite sampled averages over transformations are closely related to data augmentation and test-time ensembling \cite{ChenDobribanLee2020DataAugmentation}, and are used for non-compact groups as well: multi-crop and multi-scale evaluation average a trained network over translations and rescalings \cite{krizhevsky2012imagenet,szegedy2015going}, and
Lorentz-boost augmentation has been used to train jet taggers \cite{monaco2023lorentz}. Unlike compact Reynolds projection, such procedures depend on the chosen sampling distribution and need not define an exact projection.

Our compact-group experiments concern surrogate models for spatiotemporal dynamical systems. Learned surrogates have demonstrated strong performance across many physical domains \cite{stachenfeld2021learned,mccabe2023multiple,yang2023context,rahman2024pretraining,shen2024ups,herde2024poseidon,chen2024omniarch,sun2025towards}, and these successes have motivated large simulation benchmarks \cite{takamoto2022pdebench,audenaert2024multimodal}, including the Well \cite{ohana2024well}, which provides the datasets and pretrained models used in Section~\ref{numexp}.

Lorentz symmetry is central in particle physics and has motivated a substantial literature on symmetry-aware collider learning. Classical invariant theory implies that scalar polynomial invariants are generated by pairwise Minkowski products \cite{Weyl1939}, this viewpoint has been developed for Lorentz- and permutation-invariant particle features and invariant-polynomial machine learning \cite{GripaiosHaddadinLester2021LorentzPermutation,Haddadin2021InvariantPolynomials, villar2021scalars}. Early successful jet taggers represented jets as sets or clouds of particles, including energy-flow and particle-flow networks \cite{Komiske2019EnergyFlowNetworks}, ParticleNet \cite{Qu2020ParticleNet}, and transformer-based models such as the Particle Transformer \cite{QuLiQian2022ParticleTransformer}. Other methods incorporate relativistic structure explicitly: the Lorentz Layer introduced Lorentz-invariant combinations into top tagging \cite{ButterEtAl2018LorentzLayer}; the Lorentz Group Network builds equivariant layers from finite-dimensional Lorentz representations and tensor products \cite{BogatskiyEtAl2020LGN}; LorentzNet uses Minkowski products and four-vector message passing \cite{GongEtAl2022LorentzNet}; the Covariant Particle Transformer gives a partially Lorentz-covariant transformer for top-quark reconstruction \cite{QiuEtAl2023CPT}; and PELICAN constructs permutation-equivariant, Lorentz-invariant or covariant aggregators \cite{BogatskiyEtAl2024PELICAN}. More recent Clifford- and geometric-algebra architectures, including CGENN and L-GATr, use multivector representations to build Lorentz-aware networks for high-energy physics \cite{RuheBrandstetterForre2023CGENN,SpinnerEtAl2024LGATr,BrehmerEtAl2025AllLHC}. These works impose Lorentz structure by architecture design, whereas our focus is on post-hoc symmetrization of a given black-box predictor.

Since the Lorentz group is non-compact,  the compact Reynolds-averaging argument does not apply directly. Similar issues arise for other non-compact physical symmetries, such as dimensional rescalings. In \cite{villar2023dimensionless} the authors use the Buckingham--\(\Pi\) theorem to replace dimensional inputs by dimensionless invariant features \cite{Buckingham1914Pi} and they explain how to project functions onto the space of invariants using the Weyl unitarian trick. In Section \ref{subsec:weyl-lorentz} we show how to apply the Weyl trick to the Lorentz group. Our Weyl unitarian Lorentz average can therefore be viewed as a post-hoc invariant projection for polynomial or holomorphic predictors.

A further complication is that collider tasks often exhibit only partial Lorentz symmetry. The beam axis, detector frame, fiducial cuts, jet definitions, and preprocessing choices can introduce preferred directions or frames. Recent Lorentz-equivariant architectures address this through controlled symmetry breaking, for example, by adding reference multivectors associated with the beam or detector frame \cite{BrehmerEtAl2025AllLHC}. This is in part related to broader work showing that augmentation or equivariance can be harmful when transformed examples are not representative of the relevant data distribution \cite{LawrenceEtAl2025Augment}. Our jet-tagging experiments should therefore not be interpreted as constructing a fully Lorentz-invariant predictor. Averaging over a compact subgroup enforces invariance only under that subgroup, while averaging over a bounded boost window depends on the chosen window and is not a group projection. Instead, these experiments test whether averaging over selected physically relevant transformations improves prediction.

\section{Compact groups}
\label{sec:compact-reynolds}

\subsection{Compact group averaging and Reynolds projection}
\label{subsec:compact-reynolds}

We begin with the setting in which post-hoc symmetrization is canonical. Let \(G\) be a group acting on a vector space \(X\), and write the action as
\[
    \phi(g)x = g\cdot x,
    \qquad g\in G,\ x\in X.
\]
where
\(\phi:G\to\mathrm{GL}(X)\), with \(\mathrm{GL}(X)\) the group of invertible linear maps \(X\to X\), and \(\phi(g)\) is a matrix that specifies how elements in $X$
transform under \(g\). In particular $\phi$ is a group representation, and therefore it satisfies the transformation rules \(\phi(gh)=\phi(g)\phi(h)\) and \(\phi(g^{-1})=\phi(g)^{-1}\).

We consider another real vector space \(Y\) equipped with a representation \(\psi:G\to\mathrm{GL}(Y)\). A predictor \(f:X\to Y\) is \(G\)-equivariant if
\begin{equation} 
f(\phi(g)x)=\psi(g)f(x)
    \qquad
    \forall x\in X,\ \forall g\in G.
    \label{eq:equivariance}
\end{equation}
In the special case where \(\psi(g)=I\), this reduces to invariance: \(f(\phi(g)x)=f(x)\). In physical applications, equivariance expresses the requirement that transforming the input physical state and then predicting gives the same result as predicting first and then transforming the output.

Equivariance can be written as an invariance condition on the space of
predictors. Define the induced action
\[
    (T_g f)(x)
    :=
    \psi(g^{-1}) f(\phi(g)x).
    \label{eq:induced-action}
\]
Then \(f\) is \(G\)-equivariant if and only if
\[
    T_g f=f
    \qquad
    \forall g\in G.
\]
Thus the equivariant predictors are precisely the fixed points of the induced
action \(T_g\).

Now assume that \(G\) is compact. Then \(G\) admits a normalized Haar probability
measure \(\lambda\). That is, a measure satisfying \(\lambda(G)=1\) and the invariance property
\[
    \lambda(gA)=\lambda(Ag)=\lambda(A)
\]
for every measurable set \(A\subseteq G\) and every \(g\in G\). This allows us to define the Reynolds operator as a generalization of the concept of \emph{averaging over the group orbits}
\begin{equation}
    (Q_G f)(x)
    :=
    \int_G \psi(g^{-1}) f(\phi(g)x)\,d\lambda(g)
    =
    \int_G (T_g f)(x)\,d\lambda(g).
    \label{eq:reynolds}
\end{equation}
For a finite group, this is simply
\begin{align}
    (Q_G f)(x)
    =
    \frac{1}{|G|}
    \sum_{g\in G}
    \psi(g^{-1}) f(\phi(g)x).
    \label{eq:finite-reynolds}
\end{align}
In words, evaluate the trained model $f$ on transformed versions of the same input, transform the outputs back, and average.

The averaged predictor \(Q_G f\) is exactly \(G\)-equivariant. Indeed, for any
\(h\in G\),
\[
    T_h Q_G f
    =
    \int_G T_hT_g f\,d\lambda(g)
    =
    \int_G T_{gh} f\,d\lambda(g)
    =
    Q_G f,
\]
where the final equality follows from right invariance of Haar measure. Hence \(Q_G f\) is fixed by every \(T_h\), and therefore satisfies the equivariance condition \eqref{eq:equivariance}.

The statistical benefit of Reynolds averaging follows from its relation to orthogonal projections. Let \(\mu\) be a \(G\)-invariant probability distribution on \(X\) . Here \(L^2(X,\mu;Y)\) carries the inner product \(\langle f,h\rangle_{L^2(\mu)}=\int_X\langle f(x),h(x)\rangle_Y\,\mathrm d\mu(x)\). If we further assume that the induced group action \(T_g\) is unitary on \(L^2(X,\mu;Y)\) we have that \(\langle T_gf,T_gh\rangle_{L^2(\mu)}=\langle f,h\rangle_{L^2(\mu)}\) for all \(f,h\in L^2(X,\mu;Y)\) and \(g\in G\). We define the equivariant subspace
\[
    \mathcal H_\mu^G
    :=
    \left\{
        h\in L^2(X,\mu;Y):
        h\circ \phi=\psi \circ h
    \right\}.
\]
The \(L^2(\mu)\)-orthogonal projection onto this subspace is
\begin{equation}
    P_\mu f
    :=
    \underset{h\in\mathcal H_\mu^G}{\operatorname{argmin}}\,
    \|f-h\|_{L^2(\mu)}^2.
    \label{eq:l2min}
\end{equation}
For a compact group, if $\mu$ is $G$-invariant, the Reynolds operator coincides with this projection \cite{ElesedyZaidi2021},
\begin{equation}
    Q_Gf=P_\mu f.
    \label{eq:canonical_projection}
\end{equation}
Consequently, under squared loss with a \(G\)-equivariant target
\(f^\star\), the orthogonal decomposition gives
\[
    \mathcal R(f)-\mathcal R(Q_Gf)
    =
    \|f-Q_Gf\|_{L^2(\mu)}^2
    \geq 0.
\]
Thus under the assumptions above, Reynolds averaging cannot increase the expected risk and decreases it strictly whenever \(f\) has a nonzero non-equivariant component, equivalently whenever \(f\neq Q_Gf\) in \(L^2(\mu)\).

\subsection{Application to surrogate models}

Surrogate models estimate the state variables of a spatiotemporal dynamical system, such as a parameterized family of partial differential equations (PDEs) with fixed parameters or the observations of a physical system.
The solution to a system of $s$ state variables over a spatial domain $\mathbb{D}$ and for all times $t \ge 0$ is represented as $\mathbf{u}: \mathbb{D} \times [0, \infty) \to \mathbb{R}^s$.

The domain $\mathbb{D}$ is typically a subset of two- or three-dimensional space, such as $[0, L_x] \times [0, L_y]$ or $[0, L_x] \times [0, L_y] \times [0, L_z]$.
We employ a standard uniform discretization of the system in time and space and denote by $N_D$ the number of elements in the discretized domain $\mathbb{D}$.
A snapshot $\mathbf{u}_t: \mathbb{R} \to \mathbb{R}^{s \times N_D}$ represents the values of all $s$ state variables at all $N_D$ points in the discretized domain at a fixed time $t$.

The surrogate model $M: \mathbb{R}^{k \times s \times N_D} \to \mathbb{R}^{s \times N_D}$ is an autoregressive prediction problem that maps a sequence of $k$ snapshots $\mathbf{U}_{t_i} = [ \mathbf{u}_{t_{i-k+1}}, \ldots, \mathbf{u}_{t_{i-1}}, \mathbf{u}_{t_i} ]$ to the snapshot of the following time step such that
\begin{align} \label{uhat}
    \mathbf{u}_{t_{i+1}} \approx \hat{\mathbf{u}}_{t_{i+1}} = M(\mathbf{U}_{t_i}).
\end{align}
The goal of the model $M$ is to minimize some distance between its predictions $\hat{\mathbf{u}}_t$ and the truth $\mathbf{u}_t$.
The initial conditions at $t_0$ are given by $k$ snapshots $\mathbf{U}_{t_0}$, so that the estimated next sequence is updated according to $\hat{\mathbf{U}}_{t_{1}} = [ \mathbf{u}_{t_{-k+2}}, \ldots, \mathbf{u}_{t_0}, \hat{\mathbf{u}}_{t_{1}} ]$.
The model $M$ is applied repeatedly to generate estimates for all desired times.

Our goal is to improve already-trained models that do not inherently respect some particular symmetry of the problem, by group averaging by the use of \eqref{eq:finite-reynolds}.
In \cite{ElesedyZaidi2021} it is proven that this averaging procedure reduces the error of the predictions. The assumptions to obtain provable improvements are: 
\begin{itemize}
    \item The underlying physical problem is $G$-invariant. This often holds because governing equations satisfy symmetries. 
    \item The distribution of the training data is $G$-invariant.
\end{itemize}

Our goal is to predict the dynamics of a system with a known symmetry. We can improve a surrogate model $M$ as in \eqref{uhat} with initial sequence $\mathbf{V}_{t_0} \equiv \mathbf{U}_{t_0}$ by group averaging
\begin{align} \label{vhat}
    \hat{\mathbf{v}}_{t_{i+1}} = \frac{1}{\lvert G \rvert} \sum_{g \in G} \psi(g^{-1}) M(\phi(g) \mathbf{V}_{t_i}).
\end{align}

When the symmetry group is continuous, our approach approximates the continuous group average \eqref{eq:reynolds} using a Monte Carlo method with $n$ random samples from its elements. In the minimal case, for $n=1$, the Monte Carlo approximation reduces to
\begin{align}
    \hat{\mathbf{v}}_{t_{i+1}}^{(1)}
    = \psi(g_i^{-1})\,
      M\!\left(\phi(g_i)\mathbf{V}_{t_i}\right),
    \label{eq:one-sample-average}
\end{align}
where each $g_i$ is a randomly sampled group element of $G$.

We provide numerical examples for surrogate models in the next Section. We argue that this simple procedure provides an efficient, zero-shot method that allows us to inject physical knowledge into a model after it has been trained.

\subsection{Numerical experiments} \label{numexp}

We use the simulations and models published in ``The Well'' \cite{ohana2024well}. The project provides 15 TB of simulation data across 16 datasets, ranging from biological systems, fluid dynamics, acoustic scattering, as well as magneto-hydrodynamic simulations of extra-galactic fluids and supernova explosions.
We chose to analyze models corresponding to datasets defined on a Cartesian 2D grid with symmetric initial and boundary conditions.
For an overview of the groups used in this work, see Appendix \ref{app:groups}.
The dataset properties and state variables are summarized in Appendix \ref{expdetails}.

The \texttt{active\_matter} dataset \cite{maddu2024learning} consists of simulations of a continuum theory describing the dynamics of rod-like active particles immersed in a Stokes fluid.
The simulations were run at constant particle density and varying values for the parameters dimensionless active dipole strength and strength of particle alignment through steric interactions. The initial conditions of the quantities are plane-wave perturbations about the isotropic state.
The \texttt{gray\_scott\_reaction\_diffusion} dataset (from now on abbreviated as \texttt{gray\_scott}) contains numerical solutions of the Gray-Scott equations. They are a set of coupled reaction-diffusion equations that describe two chemical species, $A$ and $B$, whose concentrations vary in space and time. Two parameters control the ``feed'' and ``kill'' rates in the reaction. The initial conditions are either a random Fourier series or random clusters of Gaussians.
The boundary conditions are periodic for both datasets.
The underlying physical problems are sampled on a square grid and have the symmetry of the torus $\mathbb{T}^2$.
Toroidal symmetry allows all variables to translate periodically in both directions. Additionally, the problem is invariant under rotations and inversions, which is the discrete symmetry of the square $D_4$.

Additionally we analyze two datasets on a rectangular, non-square grid.
The boundary conditions are periodic in one direction and Neumann in the other.
The \texttt{rayleigh\_benard} dataset \cite{burns2020dedalus} contains simulations of convection between a hot and cold fluid layer. Initial random Gaussian noise creates density variations which causes fluid motion.
The underlying problem has the symmetry of the circle $S^1$ as well as inversions $D_2$.
The \texttt{turbulent\_radiative\_layer\_2D} dataset \cite{fielding2020multiphase} contains simulations of cold dense gas moving relative to hot dilute gas. Once turbulence is induced due to the Kelvin Helmholtz instability, mixing of the two layers occurs.
While the equations governing this process are invariant under translations and inversions, the initial conditions are not sampled from a distribution that is invariant under inversions. In all samples of the training set, the gas moves preferred in one direction. Consequently, the surrogate model does not respect this symmetry either.

The predictive power of the trained models vary greatly between the different datasets and models. Group averaging provides no improvement for the \texttt{shear\_flow} dataset because the specific initial conditions break the system's underlying symmetry, even though the theory itself and the boundary conditions are symmetric.

The CNextU-Net \cite{ronneberger2015u, liu2022convnet} was chosen in the analysis of the datasets due to its consistent performance.
The predictions $\hat{\mathbf{u}}_t$ and $\hat{\mathbf{v}}_t$ generated by \eqref{uhat} and \eqref{vhat} must be compared to the simulations $\mathbf{u}_t$. To measure how well a prediction fits the simulation, we employ the variance scaled root mean squared error (VRMSE)
\begin{align}
    \operatorname{\mathcal{L}_\text{VRMSE}}(\mathbf{u}, \mathbf{v}) = \left[
        \frac{\braket{\lvert \mathbf{u} - \mathbf{v} \rvert^2}}
        {\braket{\lvert \mathbf{v} - \braket{\mathbf{v}} \rvert^2} + \epsilon}
    \right]^{1/2},
\end{align}
where $\braket{\cdot}$ denotes the spatial mean operator, and $\epsilon = 10^{-7}$ to prevent division by zero in case $\operatorname{Var}(\mathbf{v}) = 0$. Lower $\operatorname{VRMSE}$ indicates higher agreement. $\operatorname{\mathcal{L}_\text{VRMSE}} \gtrsim 1$ signifies poor performance, as it indicates a prediction that is less accurate than a naively predicting the constant mean.

\subsection*{Results}

\begin{figure}[t]
    \small
    \centering
    \begin{subfigure}[b]{0.48\textwidth}
         \centering
         \includegraphics[width=\textwidth]{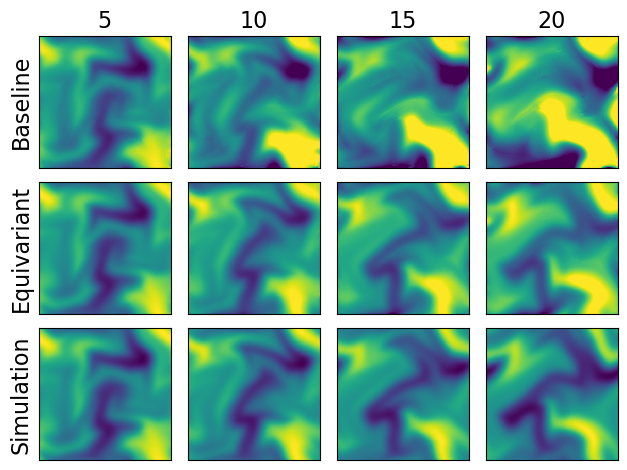}
         \caption{\texttt{active\_matter}}
         \label{active_matter_sim}
    \end{subfigure}
    \hfill
    \begin{subfigure}[b]{0.48\textwidth}
         \centering
         \includegraphics[width=\textwidth]{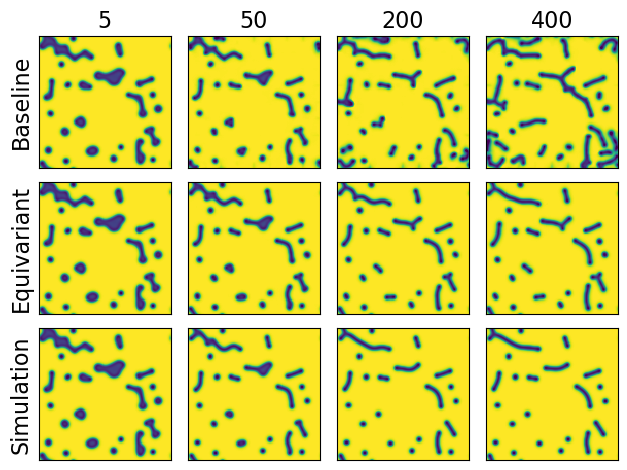}
         \caption{\texttt{gray\_scott}}
         \label{gray_scott_sim}
    \end{subfigure}
    \caption{Example comparisons of time evolutions of the concentration of two datasets.
    (\subref{active_matter_sim}) After $t=10$ steps we visually start to observe the benefits of group averaging. The equivariant model has a lower loss than the baseline at all times and is closer to the simulation (truth). At later times significant deviations between the approaches become apparent.
    (\subref{gray_scott_sim}) The equivariant model captures the features of the simulation relatively well for long times, while the baseline model develops several fictitious clusters. \label{simulations}}
    \vspace{12pt}
    \begin{subfigure}[t]{0.45\textwidth}
         \centering
         \includegraphics[width=\textwidth]{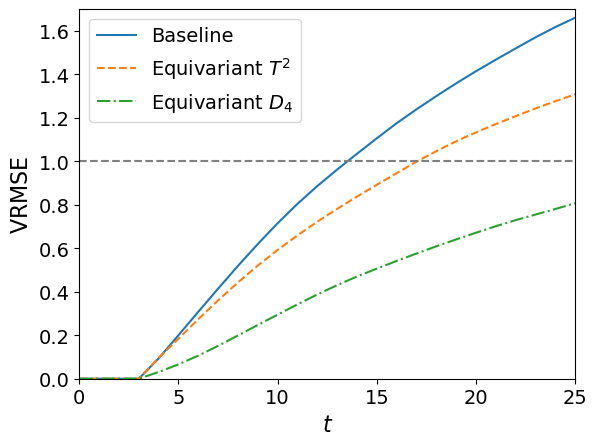}
         \caption{\texttt{active\_matter}}
         \label{active_matter_losses}
    \end{subfigure}
    \hfill
    \begin{subfigure}[t]{0.45\textwidth}
         \centering
         \includegraphics[width=\textwidth]{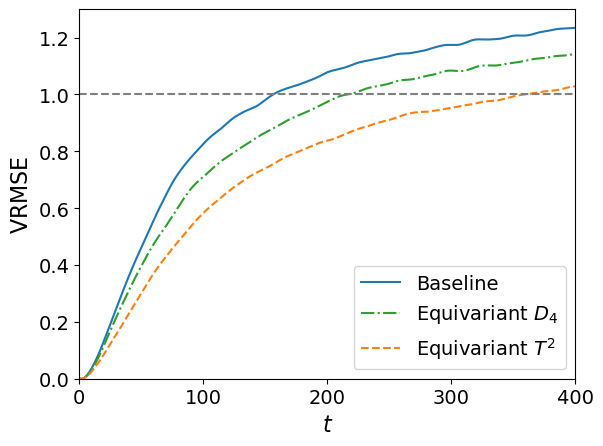}
         \caption{\texttt{gray\_scott}}
         \label{gray_scott_losses}
    \end{subfigure}
    \caption{Losses for the concentration predictions. In both cases the translations of $\mathbb{T}^2$ are performed with a single ($n=1$) random translation per time step. \label{losses}}
\end{figure}

We evaluate the performance improvement of the group averaging method by comparing the errors of the \textit{baseline} prediction
$\operatorname{\mathcal{L}_\text{VRMSE}}(\hat{\mathbf{u}}_t, \mathbf{u}_t)$
and of the \textit{equivariant} prediction
$\operatorname{\mathcal{L}_\text{VRMSE}}(\hat{\mathbf{v}}_t, \mathbf{u}_t)$.
Figure \ref{simulations} provides illustrations.
The results are summarized in Table \ref{table} and Figure \ref{losses}.

We experimented with the number of random Monte Carlo samples, $n$, used to approximate the continuous groups.
We observe significant improvements even at $n=1$ using Eq. \eqref{eq:one-sample-average}, demonstrating an enhancement over the baseline surrogate models at no additional computational cost.
Table \ref{radiative_table} shows how the group average improves as $n$ increases.

The group averaging procedure outperforms the baseline prediction on all time frames, even if we sum over only one random representative of a continuous group per time step.
At sufficiently late times, the models lose predictive power and the loss approaches 1.
The rate of loss growth and the optimal averaging group are model-dependent.

\begin{table}[t]
    \scriptsize
    \renewcommand{\arraystretch}{1.1}
    \centering
    \begin{subtable}[b]{0.48\textwidth} %
        \centering
        \begin{tabular}{@{}lccccc@{}}
            \toprule
            Model & Start & 1 & 5 & 10 & Rollout \\ \midrule
        
            Baseline & \multirow{3}{*}{10} & 0.063 & 0.386 & 0.938 & 10.81 \\
            $D_4$ & & \textbf{0.035} & \textbf{0.236} & \textbf{0.594} & \textbf{7.012} \\
            $\mathbb{T}^2$ & & 0.066 & 0.319 & 0.803 & 9.277 \\
            
            \midrule
            
            Baseline & \multirow{3}{*}{50} & 0.038 & 0.279 & 0.663 & 7.659 \\
            $D_4$ & & \textbf{0.026} & \textbf{0.193} & \textbf{0.497} & \textbf{5.821} \\
            $\mathbb{T}^2$ & & 0.038 & 0.202 & 0.507 & 5.976 \\
            
            \bottomrule
        \end{tabular}
        \caption{\texttt{active\_matter} \label{active_matter_table}}
    \end{subtable}~
    \begin{subtable}[b]{0.48\textwidth}
        \centering
        \begin{tabular}{@{}lccccc@{}}
            \toprule
            Model & Start & 1 & 10 & 50 & Rollout \\ \midrule
            
            Baseline & \multirow{3}{*}{10} & 0.008 & 0.154 & 0.742 & 1.833  \\
            $D_4$ & & \textbf{0.007} & 0.146 & 0.680 & 1.730 \\
            $\mathbb{T}^2$ & & 0.008 & \textbf{0.126} & \textbf{0.598} & \textbf{1.511} \\
            
            \midrule
            
            Baseline & \multirow{3}{*}{50} & 0.004 & 0.073 & 0.530 & 0.879 \\
            $D_4$ & & \textbf{0.003} & 0.064 & 0.453 & 0.764 \\
            $\mathbb{T}^2$ & & 0.004 & \textbf{0.051} & \textbf{0.357} & \textbf{0.608} \\
            
            \bottomrule
        \end{tabular}
        \caption{\texttt{gray\_scott} \label{gray_scott_table}}
    \end{subtable}
    \\
    \begin{subtable}[b]{0.48\textwidth}
        \centering
        \begin{tabular}{@{}lccccc@{}c}
            \toprule
            Model & Start & 1 & 10 & 20 & Rollout \\ \midrule
        
            Baseline & \multirow{3}{*}{50} & 0.065 & 0.346 & 0.627 & 4.246 \\
            $D_2$ & & \textbf{0.054} & 0.311 & 0.589 & 3.795 & \\
            $S^1$ & & 0.065 & \textbf{0.294} & \textbf{0.526} & \textbf{3.664} \\
            
            \midrule
            
            Baseline & \multirow{3}{*}{100} & 0.035 & 0.243 & 0.436 & 2.908 \\
            $D_2$ & & \textbf{0.032} & 0.216 & 0.394 & 2.586 \\
            $S^1$ & & 0.034 & \textbf{0.198} & \textbf{0.363} & \textbf{2.395} \\
            
            \bottomrule
        \end{tabular}
        \caption{\texttt{rayleigh\_benard}}
        \end{subtable}~
        \begin{subtable}[b]{0.48\textwidth}
        \centering
        \begin{tabular}{@{}lccccc@{}c}
            \toprule
            Model & Start & 1 & 5 & 10 & Rollout \\ \midrule
        
            Baseline & \multirow{3}{*}{0} & \textbf{0.142} & 0.350 & 0.579 & 7.834 \\
            $S^1_{n=1}$ & & 0.171 & 0.355 & 0.537 & 7.813 \\
            $S^1_{n=8}$ & & 0.144 & \textbf{0.306} & \textbf{0.487} & \textbf{6.911} \\
            
            \midrule
            
            Baseline & \multirow{3}{*}{50} & 0.201 & 0.499 & 0.744 & 9.128 \\
            $S^1_{n=1}$ & & 0.204 & 0.485 & 0.730 & 8.939 \\
            $S^1_{n=8}$ & & \textbf{0.174} & \textbf{0.449} & \textbf{0.722} & \textbf{8.605} \\
            
            \bottomrule
        \end{tabular}
        \caption{\texttt{turbulent\_radiative\_layer\_2D}}
        \label{radiative_table}
    \end{subtable}
    \caption{Average loss values for the datasets. The different starting positions in the trajectory and the used groups are given. The loss values are the averages of the $\operatorname{\mathcal{L}_\text{VRMSE}}$ over all trajectories and state variables in the respective test sets. The loss is evaluated at the time steps indicated above each column. The rollout loss is the sum of the first 15 errors after the initial sequence. The continuous symmetries $S^1$ and $\mathbb{T}^2$ are approximated with $n=1$ random group elements, except in (\subref{radiative_table}), as indicated.}
    \label{table}
\end{table}

\section{Non-compact groups}
\label{sec:lorentz-weyl}

For non-compact groups, the averaging construction encounters a few obstructions. We first describe the issues, then introduce the Weyl unitary trick and frame averaging as the alternatives, and finally study unit rescaling and Lorentz symmetry numerically.

\subsection{Why the compact argument fails for non-compact groups}
\label{subsec:noncompact-failure}

The first obstruction concerns the measure. If \(G\) is locally compact, then the group invariant (Haar) measure exists, but it has finite total mass if and only if \(G\) is compact. Thus, for non-compact \(G\), the formal expression
\[
    \int_G T_gf\,\dd g
\]
is not a normalized average and typically diverges. One may introduce a weight or restrict the integral to a finite window,
\begin{equation}
    Q_wf
    :=
    \int_G w(g)T_gf\,\dd g,
    \label{eq:windowed-average}
\end{equation}

but this no longer gives a canonical Reynolds projection. The weight \(w\) selects a preferred region of the group, which is shifted by the group action. Consequently, windowed averaging may be a useful numerical heuristic, but it does not generally produce an exactly \(G\)-equivariant predictor. A related localized sampling scheme is used for Lorentz data augmentation in \cite{SpinnerEtAl2025LLoCa}, where boosts are sampled from a distribution concentrated near the identity. Although this procedure is used during training rather than for post-hoc averaging, it similarly selects a preferred region of the non-compact group.

The second obstruction concerns the data distribution. Group averaging evaluates the predictor along the orbit
\[
    \{g\cdot x:g\in G\}.
\]
For a compact group, this orbit is compact, whereas for a non-compact group it may be unbounded. A trained model is typically reliable only on or near the support of its training distribution. Since the training data is finite, the training distribution is bounded, and non-compact transformations can move an input far outside this region. 

The third obstruction is statistical. As discussed in Section~\ref{subsec:compact-reynolds}, for a compact group with a \(G\)-invariant probability distribution \(\mu\), the Reynolds operator Eq. \eqref{eq:reynolds} coincides with the \(L^2(\mu)\)-orthogonal projection Eq. \eqref{eq:l2min} onto the equivariant subspace. For a non-compact group,
we may not have a probability distribution that is $G$-invariant (due to the non-compactness of orbits discussed above). We may be able to extend it to arbitrary invariant distributions, and still define \(P_\mu\) whenever that subspace is closed, but this projection depends on \(\mu\), and there is generally no group average that coincides with it. For the unit-rescaling group, Appendix~A of \cite{villar2023dimensionless} shows explicitly that the Weyl-type Reynolds projection on Laurent monomials is not, in general, the orthogonal projection for a nontrivial real data distribution. Therefore the compact-group risk-improvement identity does not extend automatically to the non-compact setting.

\subsection{The Weyl unitarian trick}
\label{subsec:weyl-trick}
In this section, we explain an analogous group averaging method for non-compact groups: the Weyl unitarian trick \cite{fulton1991representation}. It extends the group averaging construction to any group arising as a real form of a complex reductive group (for example, scalings, Lorentz, symplectic, and many other groups of relevance for physics applications) at the cost of passing through the complex numbers.

We first explain how the projection is defined for a simple example. A Laurent polynomial is a function of the form
\begin{equation} \label{eq.laurent}
    f(x)=\sum_{a\in S}c_ax^{a}
\end{equation} which allows both positive and negative integer powers and \(S\subset\mathbb Z\) finite. Consider the scaling action \(x\mapsto gx\) for \(g \in \mathbb R_{>0}\) and \(x\in\mathbb R^\times
:=\mathbb R\setminus\{0\}\). Of the functions of the form \eqref{eq.laurent}, only the constant function $\bar f(x)=c_0$ is invariant. To apply the average to this scaling group, the Weyl unitary trick first extends \(f\) from \(\mathbb R^\times\) to \(\mathbb C^\times:=\mathbb C\setminus\{0\}\) using the same Laurent-polynomial formula. The acting group \(\mathbb R_{>0}\) is likewise complexified to the multiplicative group \(\mathbb C^\times\), which contains the compact unit circle \(\mathbb T=\{z\in\mathbb C:|z|=1\}\) with normalized measure \(\lambda = d\theta/2\pi\). Averaging $f$ over this compact subgroup gives
\[ \bar f(x) = 
    \int_{\mathbb T}f(zx)\,d\lambda(z)
    =
    \sum_{a\in S}c_ax^{a}\int_{\mathbb T}z^{a}\,d\lambda(z)
    =
    c_0.
\]

Two properties of this example hold in general. The first is the configuration of subgroups. Inside \(\mathbb C^{\times}\) sit two one-dimensional real subgroups: the scaling group \(\mathbb R_{>0}\) whose symmetry we wished to impose, and the compact circle \(\mathbb T\) over which we averaged. Each has real dimension equal to the complex dimension of \(\mathbb C^{\times}\), and each has \(\mathbb C^{\times}\) as its complexification; such subgroups are the real forms of a complex group. Real forms of one group can be very different, one non-compact and another compact, and this is what the trick exploits: the symmetry is carried by one form and the averaging is carried out over the other. For connected \(G_{\mathbb C}\) the compact form is its maximal compact subgroup and is unique up to conjugacy. For the Lorentz group the two forms are \(\OO(1,3)\) and \(\OO(4)\), each of real dimension six, which become real forms of a single complex group only after the coordinate change that is described in Section~\ref{subsec:weyl-lorentz}, which identifies \(\OO_\mathbb{C}(1,3)\), with \(\OO_\mathbb{C}(4)\). The second is rigidity. Holomorphic functions (i.e. complex differentiable) are rigid, whereas smooth ones (i.e.  real infinitely differentiable) are not. A classic complex analysis theorem states that holomorphic function on a connected domain are determined by their values on a suitable slice. Passing to complex inputs buys this rigidity, and rigidity carries an invariance imposed over the compact subgroup alone to the rest of complexified group. 

We now state the general construction. Let \(G_{\mathbb C}\) be a complex reductive algebraic group with non-compact real form \(G_{\mathbb R}\) and compact real form \(K\subset G_{\mathbb C}\). Assume the input and output actions extend to algebraic representations \(\phi_{\mathbb C}:G_{\mathbb C}\to GL(X_{\mathbb C})\) and \(\psi_{\mathbb C}:G_{\mathbb C}\to GL(Y_{\mathbb C})\); and let \(\mathcal F_{\mathbb C}\) be a finite-dimensional, \(G_{\mathbb C}\)-stable space of algebraic maps \(F:X_{\mathbb C}\to Y_{\mathbb C}\) containing an extension of the real predictor \(f:X\to Y\). The induced action \(T_g\) of Section~\ref{subsec:compact-reynolds} carries over unchanged, with equivariant maps again its fixed points, and the compact \(K\) carries a normalized Haar measure \(\lambda_K\), so we may define
\begin{equation}
    (Q_KF)(u)
    :=
    \int_K
    \psi_{\mathbb C}(k^{-1})\,
    F\!\left(\phi_{\mathbb C}(k)u\right)
    \,d\lambda_K(k)
    =
    \int_K (T_kF)(u)\,d\lambda_K(k),
    \label{eq:weyl-compact-average}
\end{equation}
which is the Reynolds operator \eqref{eq:reynolds} with \(K\) in place of \(G\).

We can understand this construction in two different perspectives:

\paragraph{Group-average view.}
Similar to the compact-group average, this trick evaluates the complexified predictor at transformed inputs, transforms the outputs back, and averages against a probability measure. It is the same group averaging that has been done for the compact group, but we take a route that does the average which avoids the issues of non-compactness. 

\paragraph{Representation-theoretic view.}
The same operator has an equivalent description in the language of representations: \(Q_K\) projects onto the trivial component of the induced representation of \(K\) on \(\mathcal F_{\mathbb C}\), that is, onto the \(K\)-fixed subspace, and rigidity reads
\[
    \mathcal F_{\mathbb C}^{K}
    =
    \mathcal F_{\mathbb C}^{G_{\mathbb C}}.
\]
Retaining the trivial isotypic piece for \(K\) is therefore the same as retaining it for the whole complex group.

\paragraph{Limitations.}
The construction is narrower than compact-group averaging in two respects. First, it requires a common complex domain on which both the predictor and the Weyl unitarian projection are defined: polynomial, Laurent-polynomial, and entire holomorphic predictors qualify, whereas a network with non-analytic activations such as ReLU may not. Second, it also loses the risk guarantee. For a compact group, we have the results in Eq. \eqref{eq:canonical_projection}; for a non-compact group they do not necessarily hold. Here \(Q_K\) is canonical, closed-form, and blind to the data distribution, whereas \(P_\mu\) carries the risk interpretation but depends on \(\mu\) and has no closed form. Which of the two yields lower risk is therefore an empirical question, taken up in Section~\ref{sec:units-experiments} and Section \ref{sec:lorentz-numerical-experiments}.

\subsection{Data-dependent \texorpdfstring{\(L^2(\mu)\)}{L²(μ)} projection}
\label{sec:data-dependent-l2}

The other way to avoid averaging over a non-compact group is to define projection relative to the input distribution. Let \(\mu\) be a probability distribution on the input space \(X\), and let \(Y\) be an inner-product space. For predictors \(u,v:X\to Y\), define
\[
    \langle u,v\rangle_{\mathcal H_\mu} := \mathbb E_{x\sim\mu} \bigl[\langle u(x),v(x)\rangle_Y\bigr],
    \qquad
    \lVert u\rVert_{\mathcal H_\mu}^2 := \mathbb E_{x\sim\mu} \bigl[\lVert u(x)\rVert_Y^2\bigr],
\]
where $\mathcal H_\mu:=L^2(X,\mu;Y)$ is the space of predictors with finite \(L^2(\mu)\) norm.

Let \(\mathcal H_\mu^G\subseteq\mathcal H_\mu\) be the set of \(G\)-equivariant predictors, namely those that transform according to the prescribed group actions on \(X\) and \(Y\). We assume that \(\mathcal H_\mu^G\) is a closed linear subspace, which guarantees that every \(f\in\mathcal H_\mu\) has a unique closest equivariant predictor. This predictor is the \(L^2(\mu)\)-orthogonal projection
\begin{equation}
    P_\mu f := \operatorname*{arg\,min}_{h\in\mathcal H_\mu^G} \mathbb E_{x\sim\mu} \left[ \lVert f(x)-h(x)\rVert_Y^2 \right].
    \label{eq:l2-projection}
\end{equation}
Its residual is orthogonal to every equivariant predictor:
\[
    \langle f-P_\mu f,h\rangle_{\mathcal H_\mu}=0 \qquad \text{for every }h\in\mathcal H_\mu^G.
\]

Let \(f_\star(x)=\mathbb E[y\mid x]\) be the regression function for squared loss. If the prediction problem respects the symmetry, then \(f_\star\in\mathcal H_\mu^G\). Orthogonality therefore gives
\begin{equation}
    \lVert f-f_\star\rVert_{\mathcal H_\mu}^2 = \lVert P_\mu f-f_\star\rVert_{\mathcal H_\mu}^2 + \lVert f-P_\mu f\rVert_{\mathcal H_\mu}^2.
    \label{eq:l2-risk-decomposition}
\end{equation}
Here \(\lVert f-f_\star\rVert_{\mathcal H_\mu}^2\) is the excess squared-loss risk of \(f\). Hence \(P_\mu f\) cannot increase squared risk under \(\mu\).

For compact \(G\) and \(G\)-invariant \(\mu\), the group average is self-adjoint on \(\mathcal H_\mu\) and coincides with \(P_\mu\), so \eqref{eq:l2-risk-decomposition} recovers the compact result of Elesedy and
Zaidi~\cite{ElesedyZaidi2021}. For non-compact \(G\), Haar measure has infinite mass and \(\mu\) is not \(G\)-invariant, so an averaging operator would not be self-adjoint in \(\mathcal H_\mu\) even where one can be
defined. Averaging and projection come apart, and closedness of \(\mathcal H_\mu^G\) must be assumed rather than inherited from a bounded averaging operator.

\subsection{Frame averaging}
\label{subsec:frame-averaging-definition}

Frame averaging provides a third route to post-hoc symmetrization. Intuitively, it assigns each input a finite set of input-dependent changes of coordinates that carry it to one or more canonical representatives of its group orbit. The original predictor is evaluated at these canonicalized inputs, after which each output is transported back to the coordinates of the original input and the results are averaged. Averaging over all admissible choices removes dependence on any finite ambiguity in the canonicalization, while the consistency condition below ensures that the resulting predictor is equivariant.

Using the input and output actions \(\phi\) and \(\psi\) introduced in Section~\ref{subsec:compact-reynolds}, a finite frame assigns each \(x\in X\) a nonempty finite subset \(\mathcal F(x)\subseteq G\) satisfying
\begin{equation}
    \mathcal F(\phi(h)x)=h\mathcal F(x)
    \qquad
    \text{for every }h\in G,
    \label{eq:frame-equivariance}
\end{equation}
where \(h\mathcal F(x)=\{hg:g\in\mathcal F(x)\}\)
\cite{puny2022frameaveraginginvariantequivariant}. Given an arbitrary predictor \(f:X\to Y\), define
\begin{equation}
    (Q_{\mathcal F}f)(x)
    =
    \frac{1}{|\mathcal F(x)|}
    \sum_{g\in\mathcal F(x)}
    \psi(g)f\bigl(\phi(g^{-1})x\bigr).
    \label{eq:frame-average}
\end{equation}
Indeed, writing \(g=hu\) in the frame at \(\phi(h)x\) gives
\[
    (Q_{\mathcal F}f)(\phi(h)x)
    =
    \frac{1}{|\mathcal F(x)|}
    \sum_{u\in\mathcal F(x)}
    \psi(hu)f\bigl(\phi(u^{-1})x\bigr)
    =
    \psi(h)(Q_{\mathcal F}f)(x).
\]
Thus \(Q_{\mathcal F}f\) is exactly \(G\)-equivariant; when
\(\psi(g)=I\), it is invariant.

Because the average is finite, this construction does not require a normalized Haar measure and can be applied to a non-compact group whenever an equivariant finite frame is available. The concrete frames used below are obtained by bringing input-dependent quantities to canonical forms and averaging over any remaining finite ambiguity; their groups and regularity conditions are stated in the corresponding application sections. Although \(Q_{\mathcal F}\) fixes equivariant predictors and is therefore idempotent, it is generally not the Reynolds/Haar projection or the \(L^2(\mu)\)-orthogonal projection and hence does not automatically inherit their risk-reduction guarantees. A further limitation concerns regularity: unweighted frame averaging can turn a continuous predictor into a discontinuous equivariant one, motivating continuity preserving weighted-frame constructions \cite{dym2024equivariantframesimpossibilitycontinuous}.

\subsection{Example: Unit-rescaling group}
\label{sec:units-group}

Consider predicting the drag force on a body moving through a fluid from the fluid density \(\rho\), the flow speed \(V\), and the body diameter \(\ell\). In SI base units a configuration might read \(\rho=1.2\ \mathrm{kg\,m^{-3}}\), \(V=30\ \mathrm{m\,s^{-1}}\), \(\ell=0.1\ \mathrm{m}\), with drag force \(F=2.1\ \mathrm{N}\). Recorded in CGS units, with masses in grams and lengths in centimeters, the same configuration reads
\[
    \rho = 1.2\times10^{-3}\ \mathrm{g\,cm^{-3}},
    \quad
    V = 3.0\times10^{3}\ \mathrm{cm\,s^{-1}},
    \quad
    \ell = 10\ \mathrm{cm},
    \quad
    F = 2.1\times10^{5}\ \mathrm{dyn}.
\]
Each value is rescaled by the combination of conversion factors dictated by its physical dimensions. The physical state is unchanged; only the numbers representing it are.

A change of units is specified by one positive factor per base unit, here one each for mass, length, and time. These factors multiply under composition and are invertible, so the admissible conversions form the group \((\mathbb R_{>0})^{3}\), acting on numerical data by rescaling each coordinate by the monomial in the conversion factors determined by its dimensions. 

We ask that a predictor commute with this action: converting the inputs to CGS units, predicting, and converting the predicted force back to newtons should return the SI prediction. A model fitted to raw numerical values need not have this property, since nothing prevents it from combining coordinates in dimensionally inconsistent ways.

We now state this in general, following \cite{villar2023dimensionless}. Fix \(k\) base units and consider an input \(x=(x_1,\ldots,x_d)\in X:=(\mathbb R_{>0})^d\). The units of each coordinate \(x_i\) are encoded by a unit vector \(u_i=(u_{i1},\ldots,u_{ik})\in\mathbb Z^k\), where \(u_{ij}\) is the exponent of the \(j\)-th base unit in the units of \(x_i\). Collect these vectors as the columns of the unit matrix
\[
    U := \begin{bmatrix}
        u_1 & \cdots & u_d
    \end{bmatrix} \in\mathbb Z^{k\times d}.
\]
For \(w\in\mathbb Z^k\) and \(g\in(\mathbb R_{>0})^k\), write \(g^{w}:=\prod_{j=1}^k g_j^{w_j}\). The unit-rescaling group is
\[
    G_{\mathrm{units}}:=(\mathbb R_{>0})^k .
\]
An element \(g=(g_1,\ldots,g_k)\) rescales the \(j\)-th base unit by \(g_j\) and acts on the numerical input coordinates by
\begin{equation}
    (g\cdot x)_i
    =
    g^{-u_i}x_i
    =
    \left(\prod_{j=1}^k g_j^{-u_{ij}}\right)x_i,
    \qquad i=1,\ldots,d ,
    \label{eq:units-action}
\end{equation}
the exponent being negative because a larger unit gives a smaller numerical value. In the example above \(k=3\), the base units are \((\mathrm{kg},\mathrm{m},\mathrm{s})\), and the passage to CGS units is \(g=(10^{-3},10^{-2},1)\); the density has unit vector \(u_\rho=(1,-3,0)\), and \eqref{eq:units-action} returns the factor \(g^{-u_\rho}=(10^{-3})^{-1}(10^{-2})^{3}=10^{-3}\) used there.

Together with the assigned units vectors  \(u_1,\ldots,u_d\), the input space \(X\) is a units-typed space, which we denote by \(\mathcal Z_X\). A function $f:\mathcal{Z}_X\to \mathcal{Z}_X$ is unit-equivariant if 
\begin{equation}
    f(g\cdot x)=g\cdot f(x)
    \qquad\text{for all } g\in G_{\mathrm{units}},\ x\in X .
    \label{eq:units-equivariance}
\end{equation}
The drag force has \(v_F=(1,1,-2)\), so \(g^{-v_F}=10^{5}\), the newton-to-dyne factor. An output with \(v=0\) is dimensionless, and \eqref{eq:units-equivariance} reduces to unit invariance.

\subsubsection{Weyl projection for the unit-rescaling group}
\label{sec:weyl-units}

Although \(G_{\mathrm{units}}\) is not itself an algebraic real form, it is the identity component of \((\mathbb R^\times)^k\), the real points of a split
algebraic torus. The corresponding complex algebraic group and a compact real form are
\[
    G_{\mathbb C}=(\mathbb C^\times)^k,
    \qquad
    \mathbb T^k
    :=
    \left\{
        z\in(\mathbb C^\times)^k:
        |z_j|=1,\ j=1,\ldots,k
    \right\}.
\]
Both \(G_{\mathrm{units}}\) and \(\mathbb T^k\) are Zariski dense in
\(G_{\mathbb C}\). Consequently, a Laurent polynomial satisfies a given
units-equivariant transformation law under \(G_{\mathrm{units}}\) if and only
if its complexification satisfies the corresponding transformation law under
\(\mathbb T^k\).

Let
\[
    F_{\mathbb C}
    \in
    \mathbb C[x_1^{\pm1},\ldots,x_d^{\pm1}]
\]
be a Laurent-polynomial predictor on
\(X_{\mathbb C}:=(\mathbb C^\times)^d\). The complex torus acts by the extension
of equation~\eqref{eq:units-action}, Appendix~A of \cite{villar2023dimensionless} defines the corresponding torus
projection for invariant, dimensionless outputs. For a scalar output with units vector \(v\), the equivariant extension is
\begin{equation}
    (Q_vF_{\mathbb C})(x)
    :=
    \int_{\mathbb T^k}
    z^v F_{\mathbb C}(z\cdot x)
    \,d\lambda_{\mathbb T}(z),
    \label{eq:units-weyl-projection}
\end{equation}
where \(\lambda_{\mathbb T}\) is normalized Haar measure. The factor \(z^v\)
is the inverse output action: since the output transforms as
\(y\mapsto z^{-v}y\), transforming the output back contributes \(z^v\).

\begin{proposition}[Units Weyl unitary projection]
\label{prop:units-compact-projection}
For \(a=(a_1,\ldots,a_d)\in\mathbb Z^d\), write
\[
    x^a:=\prod_{i=1}^d x_i^{a_i}.
\]
If
\[
    F_{\mathbb C}(x)
    =
    \sum_{a\in S}c_a x^a,
    \qquad
    S\subset\mathbb Z^d
    \text{ finite},
\]
then
\[
    (Q_vF_{\mathbb C})(x)
    =
    \sum_{\substack{a\in S\\Ua=v}}c_a x^a.
\]
Thus \(Q_v\) is the projection onto the Laurent-polynomial predictors satisfying
\[
    F_{\mathbb C}(z\cdot x)=z^{-v}F_{\mathbb C}(x).
\]
In particular, \(Q_0\) retains exactly the dimensionless monomials.
\end{proposition}

\begin{proof}
Since
\[
    (z\cdot x)^a=z^{-Ua}x^a,
\]
we have
\[
    (Q_vx^a)(x)
    =
    x^a
    \int_{\mathbb T^k}
    z^{v-Ua}
    \,d\lambda_{\mathbb T}(z).
\]
The integrand is a character of \(\mathbb T^k\), whose Haar average is \(1\)
when \(Ua=v\) and \(0\) otherwise. The result follows by linearity.
\end{proof}

If \(F_{\mathbb C}\) has real coefficients, then so does \(Q_vF_{\mathbb C}\). Its restriction to \(X\) is therefore a real-valued units-equivariant predictor.

In words, \(Q_v\) creates no equivariant features; it only deletes the ones that are not. Each monomial \(x^a\) is a joint eigenvector of the scaling action, with eigenvalue \(z^{-Ua}\), so \(Q_v\) is diagonal in the monomial basis: it keeps the monomials with \(Ua=v\) and annihilates the rest. The cost is expressivity: a predictor whose dictionary contains no monomial of weight \(v\) projects to zero.

\subsubsection{Numerical experiments with the unit-rescaling group}
\label{sec:units-frame-averaging}
\label{sec:units-experiments}

We consider force regression and Reynolds-number classification tasks. The regression inputs are density \(\rho\), dynamic viscosity \(\mu\), speed \(V\), area \(A\), and length \(\ell\), with units
\(ML^{-3},ML^{-1}T^{-1},LT^{-1},L^2,L\), respectively.  Classification uses
the same variables except \(A\).  The regression target is
\[
    y_\star=\rho V^2A
    \left(0.7+0.2\,\mathrm{Re}+0.15\,\mathrm{Re}^{-1}\right),
    \qquad \mathrm{Re}:=\frac{\rho V\ell}{\mu}.
\]
Since \(\mathrm{Re}\) is dimensionless, \(y_\star\) has units of force, \(MLT^{-2}\), corresponding to the unit-weight vector \(v_F=(1,1,-2)\).

For classification, the decision boundary is \(\mathrm{Re}=1\): the regime \(\mathrm{Re}>1\) is inertia dominated, whereas \(\mathrm{Re}<1\) is viscosity dominated. We define the true logit and class probability by
\[
    s_\star(x)=2.5\left(\mathrm{Re}-\mathrm{Re}^{-1}\right),
    \qquad p_\star(x)=\operatorname{sigmoid}(s_\star(x)),
    \qquad \operatorname{sigmoid}(t):=(1+e^{-t})^{-1}.
\]

We compare the following methods:
\begin{description}
    \item[Baseline:]  following the experiments from \cite{villar2023dimensionless} we fit a ridge regression or \(\ell^2\)-penalized logistic regression on a set of features, of which 10 are units-invariant and 250 are not invariant. 
    \item[Weyl-projection:] the projection of the baseline to the space of invariant/equivariant functions using the Weyl trick in this case gives us  the restriction of the baseline model to the invariant features only. That is, zeroing out the coefficients of the non-invariant features.  
    \item[Empirical $L^2$ projection:] the $L^2$ projection of the baseline to the space of invariant/equivariant functions, obtained by minimizing Eq. \eqref{eq:l2min}, as discussed in Section \ref{sec:data-dependent-l2}.
\item[Frame average:] A change of mass, length, and time units has three
scaling factors. Each of the following triples uniquely determines a change of
units that makes its three variables equal to one:

$$
    \mathfrak S
    =\bigl\{(\rho,\mu,V),\ (\rho,\mu,\ell),\
    (\rho,V,\ell),\ (\mu,V,\ell)\bigr\}.
$$

For \(S\in\mathfrak S\), let \(U_S\) contain the powers of \(M,L,T\) in the
units of the selected variables, and define

$$
    g_S(x):=\exp\left(U_S^{-\mathsf T}(\log x)_S\right),
    \qquad
    \kappa_S(x):=g_S(x)\cdot x.
$$

Then \(\bigl(\kappa_S(x)\bigr)_S=\mathbf 1\); the remaining variables are
rescaled consistently but are not generally equal to one. Moreover,
\(g_S(q\cdot x)=g_S(x)q^{-1}\) and
\(\kappa_S(q\cdot x)=\kappa_S(x)\), so the inverse transformations
\(g_S(x)^{-1}\) form an equivariant frame.

We average the frozen baseline over these four canonicalized inputs:
\begin{equation}
(Q_{\mathrm{frame}}f)(x)
:=\frac14\sum_{S\in\mathfrak S}
g_S(x)^v f\bigl(\kappa_S(x)\bigr),
\label{eq:units-frame-average}
\end{equation}
where \(g^v:=g_M^{v_M}g_L^{v_L}g_T^{v_T}\). We use \(v=v_F\) for
regression and \(v=0\) for the classification logit, applying the sigmoid
after averaging. The factor \(g_S(x)^v\) converts the prediction back to the
original units. This is the ``Frame average \(m=4\)'' reported below. It
reuses the fitted coefficients and original RMS scales without refitting.

    \item[Invariant regression:] an invariant regression using only the invariant terms as aforementioned.
\end{description}
We train our model with two different training set, 
\begin{description}
    \item[Fixed units:] all the training data is in one fixed unit system. 
    \item[Transformed units (small shift):] the training data are transformed with random units transformations sampled from the distribution
    \begin{equation} \label{eq:train.dist.units}
    \log_{10}g_M,\log_{10}g_L,\log_{10}g_T
    \stackrel{\mathrm{iid}}{\sim}\operatorname{Unif}[-1,1].
    \end{equation}
\end{description}
 We further evaluate the performance in the following three scenarios: 
\begin{description}
    \item[Fixed units:] the test data is in one fixed unit system as above. 
    \item[Transformed units (small shift):] the test data are transformed with random units transformations sampled from Eq. \eqref{eq:train.dist.units}.
    \item[Transformed units (larger shift)] the test data are transformed with random units transformations, sampled from
    \begin{equation} \label{test.dist.units}
    \log_{10}g_M,\log_{10}g_L,\log_{10}g_T
    \stackrel{\mathrm{iid}}{\sim}\operatorname{Unif}[-2,2].
    \end{equation}
\end{description}

We also verify that the numerical group averaging produces an invariant/equivariant output. We compare the integral over a fixed grid with a Monte Carlo approximation. This, combined with further experiments studying the dependency on the number of samples are described in Appendix \ref{appendix:unit_group_other_exp}.

The regression risk is dimensionless
\[
    R=\operatorname{NMSE}
    :=\frac{\sum_i(\widehat y_i-y_i)^2}{\sum_i y_i^2},
    \qquad \operatorname{NRMSE}:=\sqrt{R}.
\]
The classification risk is the sample mean Bernoulli Kullback--Leibler (KL) divergence from the true probability \(p_i\) to the fitted probability
\(\widehat p_i\):
\[
  R_{\mathrm{KL}}:=\frac1n\sum_i
  \left[p_i\log\frac{p_i}{\widehat p_i}
  +(1-p_i)\log\frac{1-p_i}{1-\widehat p_i}\right].
\]
All risks are nonnegative and lower is better. For each test set with transformed units, the risk is first averaged over 16 held-out unit systems within each seed. The tables report mean \(\pm\) sample standard deviation over ten independently generated and fitted seeds unless stated otherwise.

For classification, the baseline model outputs a logit \(\widehat s(x)\). We apply the projection to this logit and then convert the projected logit into a probability:Regression training and validation targets contain one-percent multiplicative noise; tests use the noiseless target.

\begin{table}[ht]
\centering
\scriptsize
\setlength{\tabcolsep}{3pt}
\resizebox{\textwidth}{!}{%
\begin{tabular}{@{}lllccc@{}}
\hline
\shortstack{Task} & Training & Model
& \shortstack{Fixed units}
& \shortstack{Transformed units\\ (small shift)}
& \shortstack{Transformed units \\ (large shift)} \\
\hline
Regression
& Fixed units & Baseline
& \(0.0164\pm0.00647\) & \((1.42\pm4.27)\times10^{13}\)
& \((7.12\pm21.9)\times10^{26}\) \\
(NRMSE) & & Weyl-Projected
& \(0.0806\pm0.0641\) & \(0.0806\pm0.0637\)
& \(0.0814\pm0.0656\) \\
& & Empirical \(L^2(\mu)\)
& \(0.00936\pm0.00474\) & \(0.0100\pm0.00413\)
& \(0.0101\pm0.00379\) \\
& & Invariant regression
& \(\boldsymbol{0.00487\pm0.00125}\)
& \(\boldsymbol{0.00532\pm0.00157}\)
& \(\boldsymbol{0.00525\pm0.00163}\) \\
& & Frame average \(m=4\)
& \(2.11\pm1.26\) & \(2.02\pm1.14\)
& \(2.31\pm1.48\) \\
\cline{2-6}
& \shortstack{Transformed} & Baseline
& \(0.0325\pm0.0152\) & \(0.703\pm1.96\)
& \((1.20\pm3.55)\times10^{13}\) \\
& units & Weyl-Projected
& \(0.0398\pm0.0206\) & \(0.0390\pm0.0200\)
& \(0.0391\pm0.0200\) \\
& (small shift) & Empirical \(L^2(\mu)\)
& \(0.0402\pm0.0174\) & \(0.0409\pm0.0175\)
& \(0.0430\pm0.0243\) \\
& & Invariant regression
& \(\boldsymbol{0.00487\pm0.00125}\)
& \(\boldsymbol{0.00532\pm0.00157}\)
& \(\boldsymbol{0.00525\pm0.00163}\) \\
& & Frame average \(m=4\)
& \(0.0366\pm0.0180\) & \(0.0362\pm0.0173\)
& \(0.0374\pm0.0194\) \\
\hline
Classification
& Fixed units & Baseline
& \(0.0156\pm0.00233\) & \(10.26\pm1.38\)
& \(14.72\pm1.40\) \\
(KL) & & Weyl-Projected
& \(0.137\pm0.0514\) & \(0.138\pm0.0516\)
& \(0.138\pm0.0520\) \\
& & Empirical \(L^2(\mu)\)
& \((1.55\pm1.25)\times10^{-3}\)
& \((1.56\pm1.27)\times10^{-3}\)
& \((1.55\pm1.26)\times10^{-3}\) \\
& & Invariant regression
& \(\boldsymbol{(0.971\pm0.743)\times10^{-3}}\)
& \(\boldsymbol{(0.978\pm0.756)\times10^{-3}}\)
& \(\boldsymbol{(0.964\pm0.738)\times10^{-3}}\) \\
& & Frame average \(m=4\)
& \(0.0185\pm0.0112\) & \(0.0186\pm0.0113\)
& \(0.0186\pm0.0113\) \\
\cline{2-6}
& Transformed & Baseline
& \(0.0240\pm0.0189\) & \(0.0675\pm0.0627\)
& \(6.01\pm2.10\) \\
& units & Weyl-Projected
& \(0.0258\pm0.0194\) & \(0.0259\pm0.0194\)
& \(0.0257\pm0.0193\) \\
& (small shift) & Empirical \(L^2(\mu)\)
& \(0.0147\pm0.0142\) & \(0.0147\pm0.0141\)
& \(0.0146\pm0.0140\) \\
& & Invariant regression
& \(\boldsymbol{(0.971\pm0.743)\times10^{-3}}\)
& \(\boldsymbol{(0.978\pm0.756)\times10^{-3}}\)
& \(\boldsymbol{(0.964\pm0.738)\times10^{-3}}\) \\
& & Frame average \(m=4\)
& \(0.0247\pm0.0192\) & \(0.0248\pm0.0193\)
& \(0.0247\pm0.0192\) \\
\hline
\end{tabular}
}
\caption{Regression uses NRMSE and classification uses expected Bernoulli KL; lower is better. Here, training a new invariant model gives the best result in all cases. }
\label{tab:units-experiment-b}
\end{table}

The frame average is nearly insensitive to the test unit system, as expected from its exact equivariance. With fixed-unit training, it prevents catastrophic failures under unit shifts but can substantially reduce accuracy: the regression NRMSE remains near \(2.1\), compared with \(0.0164\) for the baseline in fixed units, while the classification KL remains near \(0.0186\) across all unit systems. With transformed-unit training, its performance improves considerably, with regression NRMSE near \(0.037\) and classification KL near \(0.025\) across all test conditions. Thus, frame averaging guarantees the correct transformation law, but its accuracy still depends on how well the frozen predictor performs on the canonicalized inputs.

The Weyl projection satisfies the prescribed units scaling law to numerical precision, so its risk is unchanged across the three unit conditions. It nevertheless increased  the in-distribution risk for both tasks. When the baseline models were tested in unit systems with transformed units, the projection reduced the regression NRMSE from \(1.56\times10^{13}\) to \(0.0821\) and the classification KL from \(10.25\) to \(0.138\).  These large gains arose because the baseline models' non-invariant terms transformed incorrectly. Using transformed units during training reduced, but did not eliminate, this sensitivity, especially for transformations out-of-distribution (such as the large shifts). 

\emph{Conclusion.}
The Weyl projection can greatly improve predictions when test examples are expressed in unit systems not seen by an already trained baseline model, because it removes dependence on the non-invariant terms.  It is not a universal accuracy improvement: it worsened the risk for the fixed units experiment. Its demonstrated benefit is its robustness to unit transformations.
Frame averaging showed the same robustness with accuracy tradeoff: it prevented the
catastrophic failures under unseen unit transformations, but it did not provide
a uniform in-distribution improvement and performed particularly poorly for the
fixed-unit regression baseline.

\subsection{Example: Lorentz group}
\label{subsec:weyl-lorentz}

We now specialize the construction of Section \ref{subsec:weyl-trick} to scalar-valued predictors that take Lorentz vectors as inputs. Invariance under \(O(1,d-1)\) means that applying the same Lorentz transformation to every input vector leaves the prediction unchanged. Unlike the unit-rescaling example, the compact real form is not contained in the real Lorentz group in the original coordinates. After Wick-rotating the time coordinate, however, the noncompact form \(O(1,d-1)\) and the compact form \(O(d)\) become real forms of the same complex orthogonal group. The projection therefore has three steps: pass to Wick coordinates, average over \(O(d)\), and restrict back to the real slice. 

Throughout, let \(d\geq 2\) and use the sign convention $\eta=\operatorname{diag}(-1,1,\ldots,1).$ For \(p=(t,x)\in\mathbb R^{1,d-1}\), with
\(x=(x_1,\ldots,x_{d-1})\), define the metric and the Minkowski invariant
\begin{equation}
    \langle p,p'\rangle_{\eta}
    :=p^{\mathsf T}\eta p'
    =-tt'+x\cdot x',
    \qquad
    s(p):=\langle p,p\rangle_{\eta}
    =-t^2+\lVert x\rVert^2.
    \label{eq:lorentz-invariant}
\end{equation}
The full Lorentz group is
\[
    \OO(1,d-1)
    =
    \{A\in \GL_d(\mathbb R):
      A^{\mathsf T}\eta A=\eta\},
\]
and its complexification is
\[
    \OO_\mathbb{C}(1,d-1)
    :=
    \{A\in \GL_d(\mathbb C):
      A^{\mathsf T}\eta A=\eta\}.
\]

Let
\[
    S=\operatorname{diag}(i,1,\ldots,1).
\]
Since \(S^{\mathsf T}S=\eta\), conjugation by \(S\) gives an isomorphism
\begin{align}
    \OO_\mathbb{C}(1,d-1)&\longrightarrow O_\mathbb{C}(d) 
    \\
    A&\longmapsto SAS^{-1}.
\end{align}
In these Wick coordinates, \(O(d)\subset O_\mathbb{C}(d)\) is a compact real form. We define the Wick embedding
\begin{align}
    W:\mathbb R^{1,d-1}&\longrightarrow\mathbb C^d
    \\
    p =(t, x_1,\ldots, x_{d-1}) &\longmapsto Sp= (it,x_1,\ldots,x_{d-1}).
    \label{eq:wick-embedding}
\end{align}
It intertwines the Lorentz action with the complex orthogonal action:
\begin{equation}
    W(\Lambda p)
    =(S\Lambda S^{-1})W(p),
    \qquad
    \Lambda\in O(1,d-1).
    \label{eq:wick-intertwining}
\end{equation}

On \(\mathbb C^d\), let
\begin{equation}
    B(u,v):=u^{\mathsf T}v,
    \qquad
    q(u):=B(u,u)
    =u_0^2+\cdots+u_{d-1}^2.
    \label{eq:complex-quadratic-form}
\end{equation}
For \(p,p'\in\mathbb R^{1,d-1}\),
\begin{equation}
    B(W(p),W(p'))
    =\langle p,p'\rangle_{\eta},
    \qquad
    q(W(p))=s(p).
    \label{eq:q-restricts-to-minkowski}
\end{equation}
Thus \(W\) converts Lorentz transformations into complex orthogonal transformations and Minkowski contractions into complex orthogonal contractions. We can therefore average in Wick coordinates and then restrict back.

\subsubsection{Lorentz invariant functions with a single four-vector input}
\label{subsec:single-lorentz-vector}

Let $f:\mathbb R^{1,d-1}\longrightarrow\mathbb R$ be a polynomial predictor, and let \(f_{\mathbb C}\) denote its coefficientwise complexification. Define the corresponding predictor in Wick coordinates by
\begin{equation}
    F(u):=f_{\mathbb C}(S^{-1}u),
    \label{eq:single-wick-extension}
\end{equation}
so that \(F(W(p))=f(p)\). Let \(K=O(d)\) act complex-linearly on \(\mathbb C^d\), and define
\begin{equation}
    (Q_KF)(u):=\int_K F(ku)\,dk,
    \label{eq:lorentz-scalar-compact-average}
\end{equation}
where \(dk\) is normalized Haar probability measure. The corresponding predictor on the Lorentz real slice is
\begin{equation}
    (Q_{\mathrm L}f)(p)
    :=(Q_KF)(W(p)).
    \label{eq:single-lorentz-projection}
\end{equation}
Haar averaging is a linear idempotent projection onto the \(K\)-invariant polynomials. Therefore \(Q_K\), and hence \(Q_{\mathrm L}\), is a linear idempotent projection on the corresponding polynomial function space. Equation \eqref{eq:lorentz-scalar-compact-average} is the scalar specialization of the average \eqref{eq:weyl-compact-average}.

\begin{proposition}[Single-vector Weyl unitary projection]
\label{prop:single-vector-lorentz-average}
Let \(f:\mathbb R^{1,d-1}\to\mathbb R\) be polynomial, and let \(F\) be
its Wick-coordinate complexification defined by
\eqref{eq:single-wick-extension}. Then there exists
\(H_F\in\mathbb C[z]\) such that
\begin{equation}
    (Q_KF)(u)=H_F(q(u)).
    \label{eq:single-vector-Hq}
\end{equation}
Consequently,
\begin{equation}
    (Q_{\mathrm L}f)(p)
    =H_F(s(p)),
    \label{eq:single-vector-lorentz-restriction}
\end{equation}
and \(Q_{\mathrm L}f\) is a real-valued,
\(O(1,d-1)\)-invariant polynomial.

\end{proposition}

\begin{proof}[Proof]
Since \(F\) is polynomial, \(Q_KF\) is polynomial. Haar invariance implies
that \(Q_KF\) is \(K\)-invariant. The compact real form
\(K=O(d)\) is Zariski dense in \(O(d,\mathbb C)\), so \(Q_KF\) is
\(O(d,\mathbb C)\)-invariant. By the first fundamental theorem for the
complex orthogonal group,
\[
    \mathbb C[u_0,\ldots,u_{d-1}]^{O(d,\mathbb C)}
    =\mathbb C[q].
\]
Hence \(Q_KF=H_F\circ q\) for some \(H_F\in\mathbb C[z]\). Equation
\eqref{eq:q-restricts-to-minkowski} then gives
\[
    (Q_{\mathrm L}f)(p)=H_F(s(p)),
\]
which is \(O(1,d-1)\)-invariant.

It remains to verify that the restriction is real-valued. Define the
antiholomorphic involution
\[
    \sigma(u):=\eta\overline{u}.
\]
Its set of fixed points is \(W(\mathbb R^{1,d-1})\). Because \(f\) has real
coefficients, the extension \eqref{eq:single-wick-extension} satisfies
\[
    F(\sigma(u))=\overline{F(u)}.
\]
Moreover,
\[
    \sigma(ku)=(\eta k\eta)\sigma(u),
    \qquad k\in O(d),
\]
and the automorphism \(k\mapsto\eta k\eta\) preserves Haar measure on \(O(d)\). Therefore
\[
    (Q_KF)(\sigma(u))
    =\overline{(Q_KF)(u)}.
\]
The averaged function is thus real-valued on the set of fixed points. 
\end{proof}

The Wick embedding is essential. For example, let \(d=2\) and \(f(t,x)=t^2\). Averaging directly over \(O(2)\) gives
\[
    \int_{O(2)} f\bigl(k(t,x)\bigr)\,dk
    =\frac{t^2+x^2}{2},
\]
which is Euclidean rather than Lorentz invariant. In Wick coordinates, however,
\[
    F(u_0,u_1)=-u_0^2,
\]
and Weyl averaging gives
\[
    (Q_{O(2)}F)(u)=-\frac12 q(u).
\]
Restricting to \(u=W(t,x)\) yields
\[
    (Q_{\mathrm L}f)(t,x)
    =-\frac12(-t^2+x^2),
\]
which depends only on the Lorentz invariant \(s(t,x)\).

\subsubsection{Lorentz invariant functions with several vector inputs}
\label{subsec:lorentz-gram}

We can extend the results above to the case of multiple input vectors. Let
\[
    P=(p_1,\ldots,p_m)\in(\mathbb R^{1,d-1})^m
\]
and write
\[
    \mathbf W(P)
    :=
    (W(p_1),\ldots,W(p_m))
    \in(\mathbb C^d)^m.
\]
The compact group \(K=O(d)\) acts diagonally by
\[
    k(u_1,\ldots,u_m)
    =(ku_1,\ldots,ku_m).
\]
For a real polynomial predictor
\[
    f:(\mathbb R^{1,d-1})^m\longrightarrow\mathbb R,
\]
let \(f_{\mathbb C}\) denote its coefficientwise complexification and
define
\[
    F(u_1,\ldots,u_m)
    :=
    f_{\mathbb C}(S^{-1}u_1,\ldots,S^{-1}u_m).
\]
Then \(F(\mathbf W(P))=f(P)\). Define
\begin{equation}
    (Q_KF)(\mathbf u)
    :=\int_K F(k\mathbf u)\,dk,
    \qquad
    (Q_{\mathrm L}f)(P)
    :=(Q_KF)(\mathbf W(P)).
    \label{eq:multivector-scalar-average}
\end{equation}

The complex bilinear Gram map is
\begin{equation}
    \Gamma:(\mathbb C^d)^m
    \longrightarrow\operatorname{Sym}_m(\mathbb C),
    \qquad
    \Gamma(\mathbf u)_{ij}
    :=B(u_i,u_j)=u_i^{\mathsf T}u_j.
    \label{eq:complex-gram}
\end{equation}
By equation~\eqref{eq:q-restricts-to-minkowski},
\begin{equation}
    \Gamma(\mathbf W(P))_{ij}
    =\langle p_i,p_j\rangle_{\eta}.
    \label{eq:lorentz-gram}
\end{equation}
The image of \(\Gamma\) is the Gram variety
\[
    \mathcal G_{m,d}
    :=
    \{G\in\operatorname{Sym}_m(\mathbb C):
      \operatorname{rank}G\leq d\}.
\]
Thus \(\mathcal G_{m,d}=\operatorname{Sym}_m(\mathbb C)\) when
\(m\leq d\), whereas for \(m>d\) the Gram entries satisfy the
corresponding determinantal relations.

\begin{proposition}[Multi-vector Weyl unitary projection]
\label{prop:multi-vector-scalar-average}
Let \(f:(\mathbb R^{1,d-1})^m\to\mathbb R\) be polynomial, and let \(F\)
be its Wick-coordinate complexification defined above. Then there exists
a polynomial
\[
    H_F\in\mathbb C[z_{ij}:1\leq i\leq j\leq m]
\]
such that
\begin{equation}
    (Q_KF)(\mathbf u)
    =H_F(\Gamma(\mathbf u)).
    \label{eq:scalar-average-factors-through-gram}
\end{equation}
Consequently,
\begin{equation}
    (Q_{\mathrm L}f)(P)
    =
    H_F\!\left(
        \bigl(\langle p_i,p_j\rangle_{\eta}\bigr)_{i,j=1}^m
    \right),
    \label{eq:scalar-average-lorentz-gram}
\end{equation}
so \(Q_{\mathrm L}f\) is real-valued and invariant under the
diagonal action of \(O(1,d-1)\). When \(m>d\), the ambient polynomial
\(H_F\) is determined only up to polynomials that vanish on
\(\mathcal G_{m,d}\).

\end{proposition}

\begin{proof}[Proof of the polynomial statement]
The average \(Q_KF\) is polynomial and is \(K\)-invariant by Haar
invariance. Since \(K=O(d)\) is Zariski dense in \(O(d,\mathbb C)\),
\(Q_KF\) is also \(O(d,\mathbb C)\)-invariant. The first fundamental
theorem for the complex orthogonal group states that invariant
polynomials on \(m\) vectors are generated by the pairwise contractions
\(B(u_i,u_j)\). Hence
\(Q_KF\) factors polynomially through \(\Gamma\), proving
\eqref{eq:scalar-average-factors-through-gram}. Equation
\eqref{eq:lorentz-gram} gives the restriction
\eqref{eq:scalar-average-lorentz-gram}. Applying the real-structure
argument from the single-vector case diagonally to
\((\mathbb C^d)^m\) shows that this restriction is real-valued.
\end{proof}

In particular, if \(f\) has total degree at most two, then its Weyl
unitary projection has the form
\begin{equation}
    (Q_{\mathrm L}f)(P)
    =
    c+\sum_{1\leq i\leq j\leq m}
    b_{ij}\langle p_i,p_j\rangle_{\eta},
    \label{eq:degree-two-lorentz-projection}
\end{equation}
for suitable real coefficients \(c\) and \(b_{ij}\).

\subsubsection{Numerical experiments for the Lorentz group}
\label{sec:lorentz-numerical-experiments}

We now test the Lorentz Weyl construction from Section~\ref{sec:lorentz-weyl}. We consider three settings. The first is a synthetic regression problem whose target is exactly Lorentz invariant by construction. This allows us to test the effect of Weyl unitarian averaging with respect to different sizes of the training data. The second is jet tagging, a realistic classification task from collider physics in which the inputs are particle four-momenta.
Analogous to improving the time-series prediction of Section~\ref{sec:compact-reynolds}, we also apply the discussed methods to a relativistic dynamic system, the driven covariant double well.

\paragraph{Synthetic experiments}
\label{sec:synthetic-experiments}

We first evaluate the Weyl projection on a controlled regression problem with an exactly Lorentz-invariant conditional mean. Each input consists of \(m=6\) four-vectors
\[
    P=(p_1,\ldots,p_m)\in(\mathbb{R}^{1,3})^m.
\]
Writing \(\eta=\operatorname{diag}(-1,1,1,1)\) and
\(G_{ij}(P)=p_i^{\mathsf T}\eta p_j\), we generate
\begin{equation}
    y(P)=\sum_{i,j=1}^{m}A_{ij}G_{ij}(P)+\epsilon,
    \qquad
    \epsilon\sim\mathcal{N}(0,0.1^2),
    \label{eq:synthetic-target}
\end{equation}
where \(A=A^{\mathsf T}\) is sampled once per seed and held fixed across training-set sizes. Since the Gram matrix is unchanged by a simultaneous Lorentz transformation of all input vectors, it is Lorentz invariant.

We consider two input distributions. In the Euclidean-isotropic setting, \(p_i\sim\mathcal{N}(0,I_4)\). In the time--space anisotropic setting, the components are independent with
\[
    t_i\sim\mathcal{N}(0,1.5^2),
    \qquad
    \mathbf{x}_i\sim\mathcal{N}(0,0.7^2 I_3).
\]

We compare four different models. 
\begin{description}
    \item[Baseline model] is a ridge regression on all coordinate monomials of degree at most two.
    \item[Spatially averaged model] post-processes the fitted predictor using a Monte Carlo \(O(3)\) average with \(N=16, 32\) samples.
    \item[Weyl-projected model] applies the Weyl \(O(4)\) projection from Section~\ref{subsec:lorentz-gram}. For degree-two polynomials, this projection eliminates the linear component and maps the quadratic component into the span of the Minkowski products \(G_{ij}\), yielding a Lorentz-invariant predictor.
    \item[Invariant baseline] is trained directly on the \(21\) distinct Gram entries that are invariant with respect to unit-rescaling group actions. 
\end{description}

\begin{figure}[ht!]
    \centering
    \includegraphics[width=1\linewidth]{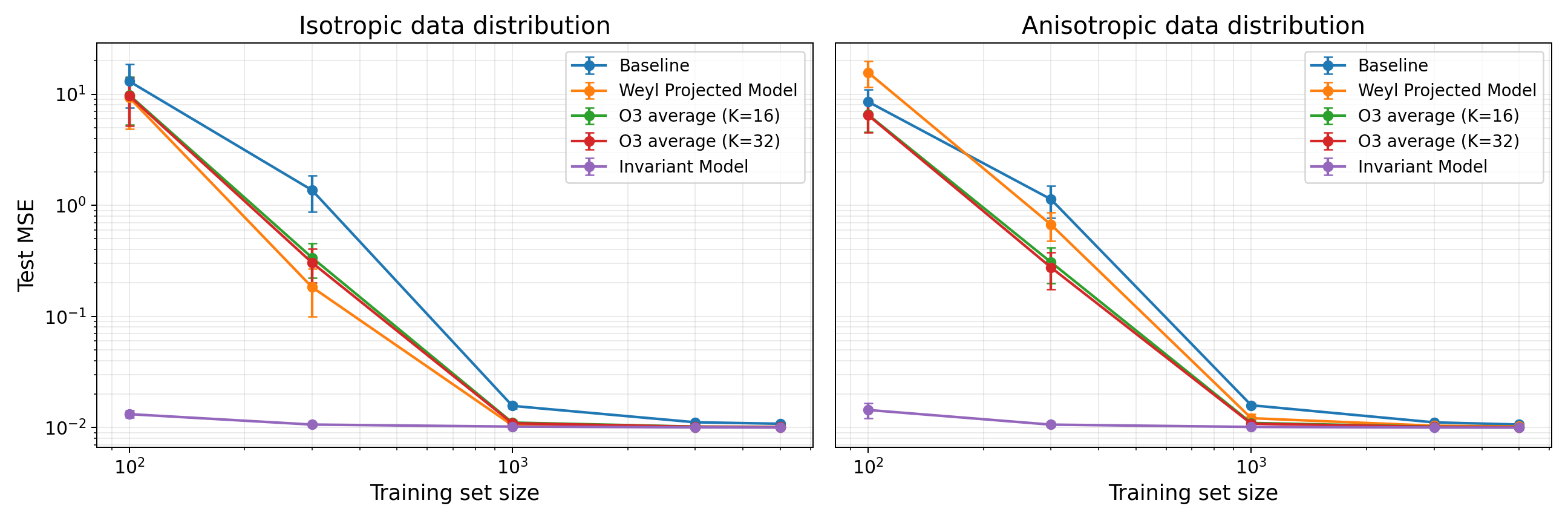}
    \caption{Test MSE for the baseline polynomial, its exact Weyl projection, the finite spatial \(O(3)\) average, and the Gram-feature invariant baseline. Both axes are logarithmic.}
    \label{fig:synthetic-mse}
\end{figure}

We use training sizes between \(100\) and \(5000\), five random seeds, and \(5000\) independent test examples per seed. Figure~\ref{fig:synthetic-mse} reports the test mean-squared error; points and error bars denote the mean and one standard deviation across seeds.

The invariant baseline is already close to the irreducible noise level \(\mathbb{E}[\epsilon^2]=0.01\) with \(100\) training examples. In the Euclidean-isotropic setting, post-training Weyl projection also substantially improves sample efficiency: at \(N_{\mathrm{train}}=300\), it reduces the MSE from \(1.37\pm0.49\) to \(0.183\pm0.083\).

This improvement is not universal. In the time--space anisotropic setting with \(N_{\mathrm{train}}=100\), Weyl projection increases the MSE from \(8.56\pm2.41\) to \(15.7\pm4.1\), whereas the \(N_{\mathrm{MC}}=32\) spatial average gives \(6.48\pm1.98\). Hence, the exact dataset distribution is important, and which specific projection used depends on the distribution. 

We also evaluate prediction variance under common Lorentz boosts. The Weyl-projected and Gram-feature models remain invariant to numerical precision, whereas the baseline and spatially averaged predictors do not. The diagnostic and complete results are reported in Appendix~\ref{appendix:unit_group_other_exp}.

\emph{Conclusion.}
With limited training data, an unconstrained model may fail to recover the symmetry of the regression function. The Weyl projection removes non-invariant components improves test error for the Euclidean-isotropic distribution, which is invariant under the compact \(O(4)\) group used in the projection. However, exact Lorentz invariance does not by itself guarantee lower prediction error: under time--space anisotropy, the Weyl projection can increase the MSE. Thus, while the projection always enforces the desired symmetry, its statistical benefit depends on the input distribution.

\paragraph{Chaotic relativistic dynamics}\label{sec:relativistic}

We now analyze a spatiotemporal chaotic relativistic dynamical system, analogous to Section~\ref{sec:compact-reynolds}.
As a toy model for a continuous, Lorentz-equivariant model we choose the driven covariant double well in $(1+1)$
dimensions.

\paragraph{System.}
The state of the particle consists of the space-time event $X=(t,x)$ and the momentum
$P=(e,p)$, with metric $\eta=\operatorname{diag}(-1,1)$ and $s(A)=\langle A,A\rangle_\eta=A^{\mathsf T}\eta A$.
All physical variables are functions of the Lorentz-invariant evolution parameter $\tau$.
The dynamics follows from the Lagrangian
\begin{equation}
  \mathcal L=\frac{\mathcal M}{2}\,s(\dot X)+\frac{\lambda}{4}\left(s(X)+b^2\right)^2
  -\kappa\,s(X)\cos\vartheta ,
  \label{eq:rel-lagrangian}
\end{equation}
where dots denote derivatives with respect to $\tau$ and $\vartheta=\Omega\tau+\vartheta_0$ is the drive phase. The equations of motion are
\begin{equation}
  \dot X=\frac{P}{\mathcal M},\qquad
  \dot P=\left[\lambda\left(s(X)+b^2\right)-2\kappa\cos\vartheta\right]X ,
  \label{eq:rel-eom}
\end{equation}
we choose $\mathcal M=b=\lambda=\Omega=1$ and drive amplitude $\kappa=0.3$. The force is a Lorentz scalar times
a Lorentz vector, so the flow of \eqref{eq:rel-eom} is exactly equivariant under the full Lorentz group
$\OO(1,1)$, generated by the boosts, parity, and time reversal
\begin{align}
    B(\xi)=\bigl(\begin{smallmatrix}\cosh\xi&\sinh\xi\\ \sinh\xi&\cosh\xi\end{smallmatrix}\bigr), \quad
    \mathcal P=\operatorname{diag}(1,-1), \quad
    \mathcal T=\operatorname{diag}(-1,1).
\end{align}
Due to the driving force, the total energy
$H=s(P)/2\mathcal M-\tfrac{\lambda}{4}(s(X)+b^2)^2+\kappa\,s(X)\cos\vartheta$ is not conserved,
$\dd H/\dd\tau=-\kappa\Omega\,s(X)\sin\vartheta$, which permits chaotic motion.

\paragraph{Data.}
We generate $N=2000$ trajectories by numerical integration and split them by trajectory into
$1400$ training, $300$ validation, and $300$ test trajectories. Since the external drive breaks
time-translation symmetry, we include the drive phase $\vartheta$ as a state variable,
encoded as $(\cos\vartheta,\sin\vartheta)$ to avoid the discontinuity at $\vartheta=2\pi$. The phase is a Lorentz
scalar and is advanced analytically, so the state is
\begin{equation}
  \mathbf u=(t,\,x,\,e,\,p,\,\cos\vartheta,\,\sin\vartheta)\in\mathbb R^6 .
  \label{eq:rel-state}
\end{equation}

We restrict to trajectories with vanishing boost charge $J=tp-xe=0$. $J$ is
conserved by \eqref{eq:rel-eom}, and the sector $J=0$, in which $X$ and $P$ are parallel, is invariant
under both the flow and $\OO(1,1)$. Trajectories in this sector remain bounded, whereas generic
trajectories with $J\neq0$ escape to infinity. All trajectories start at the
space-time origin, $X(0)=0$, with $P(0)=m_0\,(\cosh\chi_0,\sinh\chi_0)$, so that $J=0$ holds
automatically and $P(0)$ is timelike. Mass, rapidity, and phase are drawn independently with
$m_0\sim\mathrm{Rayleigh}(0.5)$, $\chi_0\sim\mathcal N(0,0.8^2)$ truncated at $|\chi_0|\le4$, and
$\vartheta_0\sim\mathcal U(0,2\pi)$. The first $\Delta\tau=30$ of each trajectory are discarded as
burn-in, and the following $\Delta\tau=150$ are recorded at intervals $\Delta\tau_s=0.1$, i.e.\
$1500$ states per trajectory. The finite-time Lyapunov exponent of the chaotic trajectories has a
median of $0.15$, a Lyapunov time of $\approx6.5$; $7\%$ of the trajectories are regular.

\paragraph{Reduced coordinates.}
On the sector $J=0$ both vectors lie on a fixed timelike ray $n=(\cosh\chi_0,\sinh\chi_0)$,
$X=\rho\,n$ and $P=\omega\,n$. The rapidity $\chi_0$ is a constant of motion, and
\begin{equation}
  \rho=\operatorname{sign}(t)\sqrt{-s(X)},\qquad
  \omega=\operatorname{sign}(e)\sqrt{-s(P)}
  \label{eq:rel-reduced}
\end{equation}
are Lorentz invariants, the position and momentum in the rest frame of the ray. In these
coordinates \eqref{eq:rel-eom} reduces to a parametrically driven Duffing double well,
\begin{equation}
  \mathcal M\ddot\rho=\left[\lambda\left(b^2-\rho^2\right)-2\kappa\cos\vartheta\right]\rho,
  \qquad \omega=\mathcal M\dot\rho ,
  \label{eq:rel-duffing}
\end{equation}
with wells at $\rho=\pm b$ separated by a barrier at $\rho=0$.
To generate trajectories, we integrate \eqref{eq:rel-duffing} and map the result to $(t,x,e,p)$.
The prediction model only ever sees the physical coordinates.

\paragraph{Surrogate model.}
As surrogate model \eqref{uhat} we use a fully connected network $f$ that maps a window of $k=8$
consecutive states \eqref{eq:rel-state} to the increment of the two Lorentz vectors. The input is the
flattened window of dimension $8\times6=48$, the network has three hidden layers of width $128$
with ReLU activations ($39\,812$ parameters), and the output is the residual
$(\Delta t,\Delta x,\Delta e,\Delta p)$. Inputs and outputs are normalized by one scalar per
Lorentz vector ($X$, $P$, $\Delta X$, $\Delta P$), computed on the training set, so that the
normalization commutes with the group action. The network is trained on all $2.1\times10^6$
windows of the training trajectories with a mean-squared loss on the normalized increments, using
AdamW with batch size $1024$ for $100$ epochs, a linear warm-up over $10^3$ steps to a learning
rate of $2\times10^{-3}$, and cosine decay to zero; we keep the checkpoint with the lowest
validation loss. Predictions are autoregressive rollouts from the first $k$
states of each test trajectory.

We test this network with three types of averages:
{\small
\begin{align} 
\textbf{Boosts average (not equivariant): }
  (Q_wf)(\mathbf U)&=\sum_{j=1}^{8}w_j\,B(\xi_j)^{-1}f\big(B(\xi_j)\,\mathbf U\big),
  &&\xi\sim\mathcal N(0,0.3^2), \label{eq:rel-boostavg}\\
  \textbf{Inversions average (equivariant): }
  (Q_{\mathcal S}f)(\mathbf U)&=\frac14\sum_{g\in\mathcal S}g\,f(g\,\mathbf U),
  &&\mathcal S=\{I,\mathcal P,\mathcal T,\mathcal P\mathcal T\}, \label{eq:rel-signavg}
\end{align}
}
and the rest-frame average $Q_{\mathcal F,w}f$ \eqref{eq:rel-restavg} below. $Q_wf$ averages over Lorentz boosts
around the frame of the input. Since the boosts are non-compact and carry no normalizable Haar
measure, the Gaussian weight is a choice and $Q_wf$ is not exactly equivariant. $Q_{\mathcal S}f$ averages over the
finite subgroup of time and space inversions; it is an exact group average.
Each group element acts on both Lorentz vectors of every state in the window and leaves the phase unchanged.

\paragraph{Rest-frame average.}
To compute a frame average as in Section \ref{subsec:frame-averaging-definition}, we apply a Lorentz boost that transforms the particle into its rest
frame. The rapidity can be computed from a short window of data: for the input window
$\mathbf U_{\tau_i}=[\mathbf u_{\tau_{i-k+1}},\dots,\mathbf u_{\tau_i}]$,
\begin{equation}
  \tanh\hat\chi_{\mathbf U}=\frac{\sum_\ell\left(t_\ell x_\ell+e_\ell p_\ell\right)}{\sum_\ell\left(t_\ell^2+e_\ell^2\right)},
  \qquad \Lambda_{\mathbf U}=B(-\hat\chi_{\mathbf U}),
  \label{eq:rel-rapidity}
\end{equation}
which is exact on the sector $J=0$, where it transforms with the input,
$\hat\chi_{B(\xi)\mathbf U}=\hat\chi_{\mathbf U}+\xi$.
Following \cite{puny2022frameaveraginginvariantequivariant}, we use the
frame $\mathcal F(\mathbf U)=\{(g\Lambda_{\mathbf U})^{-1}:g\in\mathcal S\}$, which on this sector satisfies
$\mathcal F(h\mathbf U)=h\,\mathcal F(\mathbf U)$ for all $h\in\OO(1,1)$. The frame average
\begin{equation}
  (Q_{\mathcal F}f)(\mathbf U)=\frac{1}{|\mathcal F(\mathbf U)|}\sum_{g\in\mathcal F(\mathbf U)}g\,f\big(g^{-1}\mathbf U\big)
  =\Lambda_{\mathbf U}^{-1}\,\frac14\sum_{g\in\mathcal S}g\,f\big(g\,\Lambda_{\mathbf U}\mathbf U\big)
\end{equation}
is then exactly $\OO(1,1)$-equivariant on the sector $J=0$ for any network $f$. Here we extend this procedure by
applying, in the rest frame, additional boosts in addition to the inversions.
\begin{multline}
  \textbf{Extended frame average: }(Q_{\mathcal F,w}f)(\mathbf U)=\Lambda_{\mathbf U}^{-1}\sum_{g\in\mathcal S}\sum_{j=1}^{4}\frac{w_j}{4}\,
  \big(g\,B(\zeta_j)\big)^{-1}f\big(g\,B(\zeta_j)\,\Lambda_{\mathbf U}\mathbf U\big), \\
  \qquad \zeta\sim\mathcal N(0,0.4^2), \qquad \mathcal S=\{I,\mathcal P,\mathcal T,\mathcal P\mathcal T\}
  \label{eq:rel-restavg}
\end{multline}
which requires $16$ network evaluations. Because the nodes are symmetric and
$g\,B(\zeta)\,g^{-1}=B(\pm\zeta)$ for $g\in\mathcal S$, the weighted frame is still equivariant,
and $Q_{\mathcal F,w}f$ remains exactly equivariant. Every rollout of $Q_{\mathcal F,w}f$ stays on the ray of its initial window.

The Gaussian averages in \eqref{eq:rel-boostavg} and \eqref{eq:rel-restavg} are evaluated by Gauss--Hermite
quadrature,
\begin{equation}
  \mathbb E_{\zeta\sim\mathcal N(0,\sigma^2)}\big[F(\zeta)\big]\approx\sum_{j=1}^{r}w_j\,F(\zeta_j),
  \qquad \zeta_j=\sqrt2\,\sigma\,x_j,\quad w_j=\frac{\gamma_j}{\sqrt\pi},
\end{equation}
where $x_j$ are the roots of the Hermite polynomial $H_r$ and $\gamma_j$ the corresponding weights.
We use $r=8$ for $Q_wf$ and $r=4$ for $Q_{\mathcal F,w}f$.
The widths $0.3$ and $0.4$ were selected on the validation set.

\paragraph{Error.}
The error is measured by the total deviation in the rest frame of the true trajectory. Both the
prediction and the truth are transformed by $\Lambda_{\mathrm{true}}=B(-\hat\chi_{\mathrm{true}})$, with
$\hat\chi_{\mathrm{true}}$ from \eqref{eq:rel-rapidity} applied to the whole true trajectory,
$(\tilde\rho,\tilde x)=\Lambda_{\mathrm{true}}X$ and $(\tilde\omega,\tilde p)=\Lambda_{\mathrm{true}}P$.
Then the error (or loss) $d$ is
\begin{equation}
  d(\tau)=\sqrt{\Delta\tilde\rho(\tau)^2+\Delta\tilde\omega(\tau)^2+d_\perp(\tau)^2},
  \qquad
  d_\perp(\tau)=\sqrt{\tilde x(\tau)^2+\tilde p(\tau)^2},
  \label{eq:rel-distance}
\end{equation}
where $\Delta$ denotes prediction minus truth and $d_\perp$ is evaluated on the prediction.
The first two terms measure the error within the dynamics and
$d_\perp$ the departure from the ray of the true trajectory. The distance is invariant under
a joint Lorentz transformation of prediction and truth. The reported errors are averages
over the test set at a given time $\tau$. Trajectories that erroneously escape to infinity are excluded from the test set averages.

\begin{figure}[t]
    \centering
    \begin{subfigure}[b]{0.48\textwidth}
         \centering
         \includegraphics[width=\textwidth]{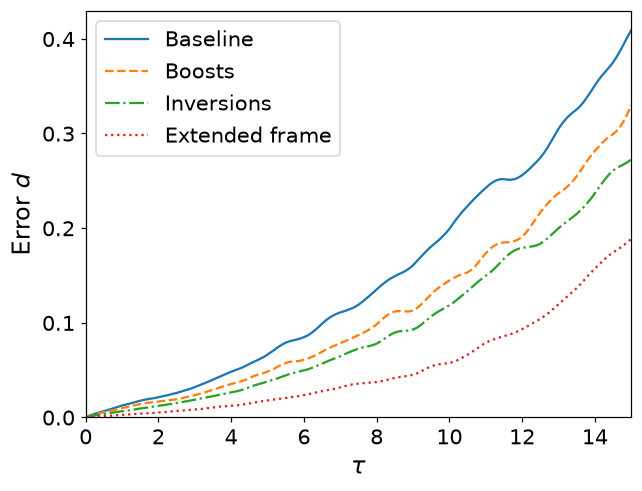}
         \caption{The rollout test-set average error $d$.}
         \label{chaos_error}
    \end{subfigure}
    \hfill
    \begin{subfigure}[b]{0.48\textwidth}
         \centering
         \includegraphics[width=\textwidth]{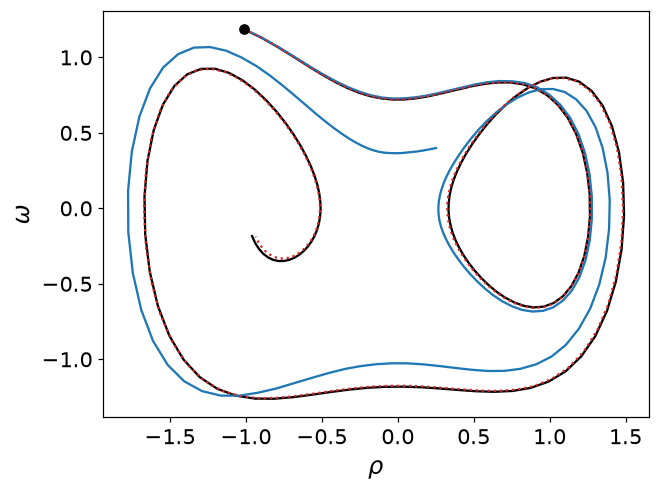}
         \caption{The phase plot of a sample trajectory.}
         \label{chaos_trajectory}
    \end{subfigure}
    \caption{
    The averaging procedures on average decrease the the rollout loss. We compare 3 different averages to the baseline prediction (the model evaluated only once). On the right, we see with an example how group averaging improves prediction: The averaged trajectory (red) stay closer to the truth (black) for longer.
    \label{fig:chaos}}
\end{figure}

Group averaging can here too significantly increase the predictive power of a model without pre-training. The results are shown in Figure \ref{fig:chaos} and Table \ref{tab:rollout}. We find that for this example a combination of the Monte-Carlo method for average boosts together with a frame-average gives the best results.

\begin{table}[t]
  \centering
  \begin{tabular}{@{}lcccc@{}}
    \toprule
    Model & 1 & 10 & 100 & Rollout \\
    \midrule
    Baseline & 0.0018 & 0.0126 & 0.2321 & 0.1497 \\
    Boosts & 0.0014 & 0.0100 & 0.1911 & 0.1201 \\
    Inversions & 0.0010 & 0.0065 & 0.1298 & 0.0786 \\
    Extended frame & \textbf{0.0003} & \textbf{0.0027} & \textbf{0.0674} & \textbf{0.0313} \\
    \bottomrule
  \end{tabular}
    \caption{Average invariant loss values $d$ for the test-set ($N=300$ test trajectories) after $n=1,10,100$ autoregressive steps with $\Delta\tau_s=0.1$ per step.
    The rollout loss is the sum of the first 15 errors.}
    \label{tab:rollout}
\end{table}

\paragraph{Jet tagging}
\label{subsec:jet-tagging}

We next study quark--gluon jet tagging, a standard classification problem in collider physics.  A jet is represented as an unordered collection of constituent four-momenta,
\begin{equation}
    X=\{p_i\}_{i=1}^{n_X},
    \qquad
    p_i=(E_i,p_{x,i},p_{y,i},p_{z,i}).
    \label{eq:jet-input}
\end{equation}
We use the two-class Pythia8 quark--gluon sample distributed with EnergyFlow
\cite{Komiske2019QGDataset,Komiske2019EnergyFlowNetworks}, with quark jets treated as signal and gluon jets as background.  The split contains \(1.6\times 10^6\) training jets, \(2\times 10^5\) validation jets, and \(2\times 10^5\) test jets, with equal numbers of the two classes in each split. 

We evaluate two complementary classifiers. The first is ParticleNet-Lite
\cite{Qu2020ParticleNet}, a realistic high-capacity jet tagger. The second is a degree-three polynomial classifier whose logit is a polynomial in the Cartesian components \((E_i,p_{x,i},p_{y,i},p_{z,i})\) of the constituent momenta, built from the permutation-invariant aggregates \(N=n\), \(A=\sum_i p_i\in\mathbb{R}^{4}\) and \(B=\sum_i p_i p_i^{\top}\in\mathbb{R}^{4\times4}\), which supply \(D=15\) features once the symmetry of \(B\) is used, and fitted by \(\ell_2\)-penalized logistic regression. It serves two purposes. Because it retains explicit dependence on the azimuthal coordinate, it isolates the effect of post-training \(SO(2)\) averaging on a predictor whose fitted coefficients are not already azimuthally invariant. And because its logit is polynomial in the four-momenta, it lies in the function class on which the Weyl projection of Section~\ref{subsec:weyl-lorentz} is defined.

Across the two models, we consider the following post-hoc averaging procedures:
\begin{description}
\item[Beam-axis \(SO(2)\) averaging.]
We average over rotations about the beam axis. This is a compact-group average aligned with the azimuthal symmetry of the proton--proton collision sample.

\item[\(SO(3)\) averaging.]
We average over spatial rotations while holding the laboratory beam axis fixed. Because these transformations do not preserve the saved-jet distribution, this provides a deliberate symmetry-mismatch control.

\item[Bounded boost averaging.]
We separately average over boosts parallel and transverse to the beam
axis, with rapidities drawn from a fixed bounded law,
\(\rho\sim\mathcal N(0,0.3^{2})\) rejection-truncated to
\(|\rho|\leq0.3\).  Since boosts form a non-compact group, these are
windowed test-time averages rather than Reynolds projections.

\item[Weyl-unitarian projection.] 
For degree-three polynomial classifier, we use the formula derived and applied to the polynomial to act as the classifier. 

\item[Collider-aligned frame averaging.]
We also apply the frame average from Section~\ref{subsec:frame-averaging-definition} to ParticleNet-Lite. Let

$$
    P=(P_0,P_x,P_y,P_z):=\sum_i p_i,
    \qquad
    p_T:=\sqrt{P_x^2+P_y^2},
    \qquad
    m_T:=\sqrt{P_0^2-P_z^2}.
$$

We remove the jet azimuth and longitudinal rapidity through
\begin{equation}
\phi_J:=\operatorname{atan2}(P_y,P_x),
\qquad
y_J:=\operatorname{atanh}\left(\frac{P_z}{P_0}\right),
\qquad
q_i:=B_z(-y_J)R_z(-\phi_J)p_i,
\label{eq:collider-canonicalization}
\end{equation}
so that \(\sum_iq_i=(m_T,p_T,0,0)\). We then average over the residual reflection sets

$$
    \mathcal R_1=\{I\},\qquad
    \mathcal R_2=\{I,r_y\},\qquad
    \mathcal R_4=\{I,r_y,r_z,r_yr_z\},
$$

where \(r_y\) and \(r_z\) reverse the corresponding momentum components. For a scalar predictor \(f\), the resulting \(K\)-view predictor is
\begin{equation}
(Q_{\mathrm{coll},K}f)(X)
:=\frac{1}{K}\sum_{r\in\mathcal R_K}
f\bigl({r q_i}_{i=1}^{n_X}\bigr).
\label{eq:collider-frame-average}
\end{equation}
The construction enforces invariance under rotations about and boosts along the beam axis, together with the selected reflections, and is well defined for \(p_T>0\) and \(\lvert P_z/P_0\rvert<1\).

\end{description}

For each probability-valued classifier \(f\), we form
\begin{equation}
    (Q_{\mathcal G,M}f)(X)
    =
    \frac{1}{M}
    \sum_{\ell=0}^{M-1} f(g_\ell X).
    \label{eq:jet-finite-average}
\end{equation}
Each transformation acts on every constituent four-momentum in the jet before all derived kinematic features are recomputed. For \(SO(2)\), we use the deterministic angles \(\theta_\ell=2\pi\ell/M\).  For \(SO(3)\), we use a fixed list of Haar-sampled proper rotations while holding the laboratory beam axis fixed. The ParticleNet-Lite results below use \(M=128\), average the two class probabilities, and use the same transformation list for every jet and every trained model.

To resolve changes hidden by aggregate accuracy, let \(N_{-+}\) be the number of test jets whose prediction changes from incorrect before averaging to correct after averaging, and let \(N_{+-}\) count changes in the opposite direction.  We report
\begin{equation}
    \Delta N_{\mathrm{correct}}
    :=
    N_{-+}-N_{+-},
    \label{eq:jet-net-corrected}
\end{equation}
which is exactly the change in the number of correctly classified test jets. We also report the background rejection \(\operatorname{Rej}_{\alpha}=1/\epsilon_B\) at quark efficiencies
\(\alpha=50\%\) and \(30\%\), where \(\epsilon_B\) is the gluon acceptance at the threshold chosen to obtain the stated quark efficiency.

\begin{table}[ht!]
    \centering
    \resizebox{\linewidth}{!}{%
    \begin{tabular}{lrrrrrr}
        \toprule
        Predictor
        & Accuracy
        & \(N_{-+}\)
        & \(N_{+-}\)
        & \(\Delta N_{\mathrm{correct}}\)
        & \(\operatorname{Rej}_{50\%}\)
        & \(\operatorname{Rej}_{30\%}\) \\
        \midrule
        Baseline (ParticleNet-Lite)
        & \(0.825031 \pm 0.000293\)
        & --- & --- & ---
        & \(33.294 \pm 0.229\)
        & \(84.064 \pm 0.820\) \\

        Collider-aligned frame average (\(K=4\))
        & \(0.825078 \pm 0.000308\)
        & \(2026.0 \pm 38.7\)
        & \(2016.6 \pm 48.2\)
        & \(+9.4 \pm 36.7\)
        & \(33.408 \pm 0.218\)
        & \(84.597 \pm 0.503\) \\

        \(SO(2)\) average
        & \(0.825032 \pm 0.000295\)
        & \(0.33 \pm 0.50\)
        & \(0.22 \pm 0.44\)
        & \(+0.11 \pm 0.78\)
        & \(33.294 \pm 0.229\)
        & \(84.064 \pm 0.820\) \\

        \(SO(3)\) average
        & \(0.799649 \pm 0.001650\)
        & \(5804.3 \pm 121.5\)
        & \(10880.8 \pm 408.2\)
        & \(-5076.4 \pm 313.6\)
        & \(28.964 \pm 0.252\)
        & \(74.424 \pm 1.040\) \\

        Longitudinal boost average
        & \(0.825184 \pm 0.000330\)
        & \(287.7 \pm 16.0\)
        & \(257.1 \pm 12.9\)
        & \(+30.6 \pm 22.8\)
        & \(33.350 \pm 0.200\)
        & \(84.007 \pm 0.679\) \\

        Transverse boost average
        & \(0.824858 \pm 0.000412\)
        & \(951.3 \pm 47.5\)
        & \(986.0 \pm 46.1\)
        & \(-34.7 \pm 30.5\)
        & \(33.259 \pm 0.177\)
        & \(83.587 \pm 0.776\) \\
        \bottomrule
    \end{tabular}}
    \caption{ParticleNet-Lite performance before and after spatial-rotation, bounded-boost, and input-dependent collider-aligned frame averaging. Entries are the mean \(\pm\) sample standard deviation over nine independently trained models evaluated on the same 200,000-jet test set. The spatial-rotation and bounded-boost entries use 128 frames and average class probabilities. The collider-aligned entry uses four deterministic canonical views and averages pre-softmax binary log-odds. The boost rapidities are sampled from a rejection-truncated Gaussian \(\rho\sim\mathcal{N}(0,0.3^2)\), with \(|\rho|\leq0.3\). Here \(N_{-+}\) counts incorrect-to-correct changes and \(N_{+-}\) counts correct-to-incorrect changes.}
    \label{tab:jet-tagging-results}
\end{table}

\(SO(2)\) averaging leaves the ParticleNet predictions unchanged to numerical precision, as a result of the construction of the model that extract numerical values of a jet making it invariant with respect to $SO(2)$ changes.  Fixed-lab \(SO(3)\) averaging, however, changes many decisions in both directions.  This shows that the exact Lorentz symmetry is not present in the Quark-gluon dataset as the particles are recorded with a fixed beam axis. The Collider-aligned frame average is the method which showed consistent improvements across all seeds.

To probe the finite-sample effect of beam-axis averaging and also attempt to apply an exact Weyl unitarian trick, we also train the degree-three polynomial classifier. For each size and each of ten seeds, all post-hoc methods are applied to the best checkpoint. 

\begin{figure}[ht!]
    \centering
    \includegraphics[width=\linewidth]{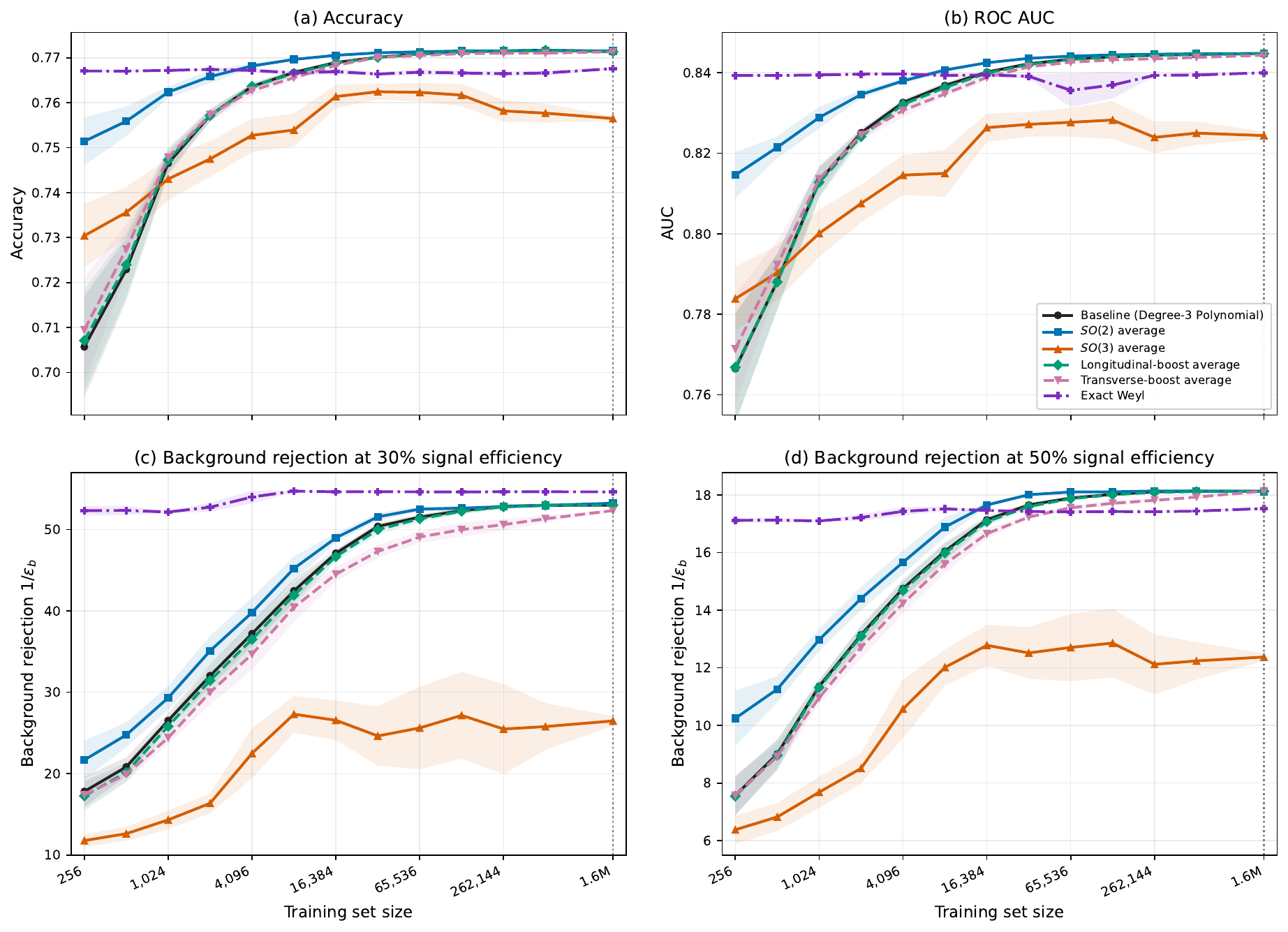}
    \caption{ROC AUC and background rejection at quark efficiencies of \(30\%\) and \(50\%\) for the degree-three polynomial classifier. }
    \label{fig:jet-tagging-results}
\end{figure}

Figure~\ref{fig:jet-tagging-results} shows that beam-axis \(SO(2)\) averaging improves AUC throughout the training-size sweep, with the largest gain in the low-data regime and a diminishing advantage as the training set grows. This is consistent with compact-group projection.  Averaging removes non-invariant components that become less important as the baseline learner sees more data \cite{ElesedyZaidi2021}. 

The Exact Weyl projection supplies a stronger low-data inductive bias. The saved-jet distribution is not strictly Lorentz invariant, for the beam-axis reason described above, yet restricting the classifier to the fixed \(18\)-dimensional invariant subspace already saturates its achievable performance within a few hundred examples. That subspace is spanned by the products of total degree at most three in \(N\), the Minkowski traces \(\operatorname{tr}\hat{B}\), \(\operatorname{tr}\hat{B}^{2}\), \(\operatorname{tr}\hat{B}^{3}\) of \(\hat{B}^{\mu}{}_{\nu}=\eta^{\mu\rho}B_{\rho\nu}\), and the contractions \(A_\mu A^\mu\) and \(A_\mu B^{\mu\nu} A_\nu\). The projection collapses the \(816\)-coefficient hypothesis class onto these
\(18\) directions, and the surviving coefficients converge almost immediately: the projected discriminant direction at \(256\) examples lies within one degree of its \(1.6\times10^{6}\) counterpart, while its norm grows by a factor of \(23\). Because a positive rescaling leaves the ordering of scores unchanged, AUC and background rejection are flat across the entire sweep. The projected model is therefore the strongest predictor in the low-data regime, but it stops improving, but it stops improving: the unprojected baseline overtakes it in AUC and at
\(50\%\) quark efficiency between \(8{,}192\) and \(16{,}384\) examples, while at \(30\%\) it stays at or above the baseline across the sweep. The ROC curves cross, so the projection remains competitive at the high-purity working points.

Overall, performance follows the alignment between the imposed transformation and the collider setup. Beam-axis \(SO(2)\) preserves that setup, whereas fixed-laboratory \(SO(3)\) averaging is detrimental because arbitrary spatial rotations respect neither the beam direction nor the selected-jet distribution. The bounded longitudinal-boost average stays close to the raw model, while transverse averaging is generally mildly detrimental. We therefore read the boost curves only as bounded-window sensitivity checks rather than canonical projections over the noncompact Lorentz group; their uncertainty intervals additionally condition on a single fixed transformation bank.

\section{Discussion and conclusion}
\label{sec:discussion-conclusion}

This work asks how a known physical symmetry can be imposed on an already trained predictor, and when doing so improves prediction. For compact groups, these objectives are closely connected. As shown in Section~\ref{subsec:compact-reynolds}, normalized Haar averaging defines the Reynolds projection onto the invariant or equivariant subspace. When the input data distribution is invariant, the induced group action is unitary, and the squared-loss regression function is equivariant, this operator coincides with the \(L^2(\mu)\)-orthogonal projection in Eq.~\eqref{eq:canonical_projection} and therefore cannot increase population squared risk. The surrogate-model experiments of Section~\ref{numexp} are consistent with this theory: averaging generally reduced prediction errors when the implemented transformations were genuine symmetries of the complete learning problem.

For non-compact groups, this correspondence breaks down. There is no normalized Haar probability measure, so the compact Reynolds construction cannot be applied directly. More fundamentally, the algebraic projection that enforces a transformation law need not coincide with the \(L^2(\mu)\) projection that minimizes error under a specified data distribution. The data-dependent projection of Section~\ref{sec:data-dependent-l2} retains the latter statistical interpretation, but its implementation requires to solve an optimization problem akin to training an equivariant model. The Weyl construction of Section~\ref{subsec:weyl-trick} instead provides a distribution-independent algebraic projection for the polynomial, Laurent-polynomial, and suitably holomorphic function classes covered by the construction. It enforces the desired symmetry exactly, but it does not inherit the compact-group risk guarantee.

Frame averaging, in Section~\ref{subsec:frame-averaging-definition}, is a third route. It replaces the integral over the group with a finite average over an input-dependent set of transformations, so it requires neither a normalized
measure nor an analytic predictor and applies to an arbitrary black-box model. Like the Weyl construction it is exact, in that the averaged predictor satisfies the prescribed transformation law identically, and like the Weyl construction it is in general neither the Haar projection nor the \(L^2(\mu)\) projection, so it carries no risk guarantee either.

The non-compact-group experiments give four main empirical conclusions. First, the Weyl projection did not consistently improve in-distribution accuracy, but it produced substantial gains with limited training data and when test inputs were expressed in unit systems not seen during training. Second, frame averaging gives the same exactness without the function-class restriction: it is units-equivariant by construction and applies to an arbitrary black-box predictor, and it prevented the catastrophic failures the baseline suffers under unseen unit transformations. Its in-distribution cost was severe for the fixed-unit-trained regression. Third, the empirical \(L^2(\mu)\) projection improved on the baseline under most tested unit shift and in-distribution task and training-regime combinations, illustrating the benefit of adapting the projection to the data distribution; it was also the least stable of the three, degrading sharply at small training-set sizes (Appendix C). Fourth, when retraining was possible, models trained directly using symmetry-adapted features matched or outperformed the post-hoc projections in our controlled unit-rescaling and Lorentz experiments. These comparisons position post-hoc symmetrization as a complement to symmetry-aware training, particularly when retraining or architectural modification is impractical.

Several questions remain open. Theoretically, it would be useful to characterize the distributions and predictor classes for which the Weyl projection reduces risk, and to quantify its discrepancy from the \(L^2(\mu)\) projection in terms of the geometry of the data distribution. Computationally, extending the construction beyond polynomial or holomorphically extendable predictors would be necessary for its direct application to modern black-box neural networks. %

\section*{Acknowledgements}

SV and WZ were partially funded by  NSF DMS 2031985. SV is also funded by NSF CAREER 2339682,  NSF
BSF 2430292, and a Sloan Fellowship. We thank Lara Knezevic for drawing our attention to related work on twirling in quantum machine learning.

\section*{Data availability statement}
The data that support the findings of this study are openly available. The quark--gluon jet dataset is available at
\url{https://doi.org/10.5281/zenodo.3164691}. The Well dataset used in section~\ref{sec:compact-reynolds} is part of the Well collection~\cite{ohana2024well} and is available at \url{https://polymathic-ai.org/the_well/}. The synthetic unit-rescaling and Lorentz datasets, together with the code used to generate them and to reproduce all experiments reported here, are available at \url{https://github.com/wenjie271/Post-hoc-group-averaging}.

\section*{Conflicts of interest statement}
The authors have no conflicts of interest. 

\section*{Ethics statement}
Not applicable.

\printbibliography

\appendix

\section{Groups} \label{app:groups}

\subsection{Dihedral groups}

For auto-regressive prediction problems, we are interested in datasets on a square or rectangular Cartesian grid. Datasets defined on a square grid with appropriate initial and boundary conditions have the largest possible symmetry.

 The symmetry group of the square is $D_4$ and contains $\lvert D_4 \rvert = 8$ elements. For datasets defined on a rectangle, the biggest symmetry group is $D_2$, containing $\lvert D_2 \rvert = 4$ elements. Inversions correspond to $D_1 \cong C_2 \cong Z_2$ with $\lvert D_1 \rvert = 2$ elements.

Generally, the dihedral groups $D_n$ refer to the symmetry groups of the $n$-gon, groups of order $2n$. We can think of these as $n$ rotational symmetries and $n$ reflection symmetries.

The group representation of $D_4$ is
\begin{align}
    D_4 = \braket{r, i \vert r^4 = i^2 = (i r)^2 = e},
\end{align}
where its members can be thought of as rotations and inversions.
One representation is given by the matrices
\begin{align}
\begin{split} \label{matrep}
    R = \begin{pmatrix}
        0 & -1 \\
        1 & 0
    \end{pmatrix}, \quad
    I = \begin{pmatrix}
        -1 & 0 \\
        0 & 1
    \end{pmatrix},
\end{split}
\end{align}
and the $\lvert D_4 \rvert = 8$ group elements can be constructed by
\begin{align}
     \id, R, R^2, R^3, I, I R, I R^2, I R^3.
\end{align}
$D_2$ can be constructed as above using only even powers of $R$, and $D_1$ by using only the inversion.

\subsection{Circle groups}

Translation invariant problems with periodic boundary conditions obey the symmetry of the circle.
The circle group is an example of a compact continuous Lie group. It is known by several names, such as $T \cong S^1 \cong U(1) \cong SO(2)$. The irreducible representations are one-dimensional maps to the unit circle.
If the problem is translation invariant and periodic in two directions, it is defined on a 2-torus. 
The torus group $\mathbb{T}^2$ is the continuous Abelian group formed from the direct product of two circle groups $\mathbb{T}^2 \cong S^1 \times S^1$.

Since numerical simulation data is always given on a discrete grid of dimension $N \times M$, we approximate the continuous groups with discrete cyclic groups $C_N \cong \mathbb{Z}/\mathbb{Z}_N$, that is $S^1$ with $C_N$ and $\mathbb{T}^2$ with $C_N \times C_M$.

\newpage

\section{Experimental details} \label{expdetails}

\subsection{Compact group averaging}

``The Well''\footnote{\url{https://polymathic-ai.org/the_well/}} \cite{ohana2024well} is published under \href{https://creativecommons.org/licenses/by/4.0/}{CC BY 4.0}.
Each test set contains $0.1 \times N$ of the trajectories. We excluded 9 trajectories from the \texttt{gray\_scott} dataset that converge to a steady state in our analysis.

\begin{table}[ht]
    \centering
    \renewcommand{\arraystretch}{1.1}
    \caption{Dataset specifications}
    \label{datasets}
    \vspace{10pt}
    \begin{subtable}[b]{\textwidth}
        \centering
        \begin{tabular}{@{}lccccc@{}}
            \toprule
            Dataset & Size (GB) & $N_D$ & $T$ & $N$ & $s$ \\ \midrule
            \texttt{active\_matter} & 51.3 & $256 \times 256$ & 81 & 360 & 11 \\
            \texttt{gray\_scott} & 153.8 & $128 \times 128$ & 1001 & 1200 & 2 \\
            \texttt{rayleigh\_benard} & 342 & $512 \times 128$ & 200 & 1750 & 4 \\
            \texttt{turbulent\_radiative\_layer\_2D} & 6.9 & $384 \times 128$ & 101 & 90 & 4 \\
            \bottomrule
        \end{tabular}
        \caption{Total size, resolution $N_D$, number of time steps per trajectory $T$, number of trajectories in the data set $N$, number of state variables $s$.}
        \label{data_specs}
    \end{subtable} \\
    \vspace{10pt}
    \begin{subtable}[b]{\textwidth}
        \centering
        \begin{tabular}{@{}llcc@{}}
            \toprule
            Dataset & Quantity & Type & Components \\ \midrule
            \multirow{4}{*}{\texttt{active\_matter}}
            & Concentration & Scalar & 1 \\
            & Velocity & Vector & 2 \\
            & Orientation & Tensor & 4 \\
            & Rate-of-strain & Tensor & 4 \\\midrule
            \multirow{2}{*}{\texttt{gray\_scott}}
            & Concentration $A$ & Scalar & 1 \\
            & Concentration $B$ & Scalar & 1 \\ \midrule
            \multirow{3}{*}{\texttt{rayleigh\_benard}}
            & Buoyancy & Scalar & 1 \\
            & Pressure & Scalar & 1 \\
            & Velocity & Vector & 2 \\ \midrule
            \multirow{3}{*}{\texttt{turbulent\_radiative\_layer\_2D}}
            & Buoyancy & Scalar & 1 \\
            & Pressure & Scalar & 1 \\
            & Velocity & Vector & 2 \\
            \bottomrule
        \end{tabular}
        \caption{The state variables of the datasets.}
        \label{statevariables}
    \end{subtable}
\end{table}

\subsection{Jet-tagging dataset}\label{app:jets}

The quark--gluon jet dataset~\cite{Komiske2019EnergyFlowNetworks} is published under CC~BY~4.0.\footnote{\url{https://doi.org/10.5281/zenodo.3164691}} The sample was generated with \textsc{Pythia}~8.226 at $\sqrt{s} = 14\,\mathrm{TeV}$, with quark jets produced via $qg \to Z(\to \nu\bar\nu)\,q$ and gluon jets via $q\bar q \to Z(\to \nu\bar\nu)\,g$; no detector simulation or pileup is included. Final-state particles are clustered with the anti-$k_T$ algorithm ($R = 0.4$, \textsc{FastJet}~3.3.0), and jets with $p_T^{\mathrm{jet}} \in [500, 550]\,\mathrm{GeV}$ and $|y^{\mathrm{jet}}| < 1.7$ are retained. We use the default sample, which excludes charm and bottom quark jets at the level of the hard process. The dataset contains $2 \times 10^6$ jets with equal numbers of the two classes; we use $1.6 \times 10^6$ jets for training, $2 \times 10^5$ for validation, and $2 \times 10^5$ for testing, preserving the class balance in each split. Each jet constituent is stored as $(p_T, y, \phi, \mathrm{PDG\ id})$, from which we reconstruct the Cartesian four-momenta $(E_i, p_{x,i}, p_{y,i}, p_{z,i})$ of equation~\eqref{eq:jet-input}.

\begin{table}[h]
\centering
\begin{tabular}{ll}
\hline
Generator & \textsc{Pythia}~8.226, $\sqrt{s} = 14\,\mathrm{TeV}$ \\
Processes & $qg \to Z(\to\nu\bar\nu)q$ (quark), \ $q\bar q \to Z(\to\nu\bar\nu)g$ (gluon) \\
Clustering & anti-$k_T$, $R = 0.4$ (\textsc{FastJet}~3.3.0) \\
Selection & $p_T^{\mathrm{jet}} \in [500,550]\,\mathrm{GeV}$, $|y^{\mathrm{jet}}| < 1.7$ \\
Constituent features & $(p_T, y, \phi, \mathrm{PDG\ id})$ \\
Split (train/val/test) & $1.6\times 10^6 \,/\, 2\times 10^5 \,/\, 2\times 10^5$ \\
Classes & quark (signal) vs.\ gluon (background), balanced \\
\hline
\end{tabular}
\caption{Specifications of the quark--gluon jet dataset~\cite{Komiske2019EnergyFlowNetworks}.}
\label{tab:jet-dataset}
\end{table}

\paragraph{Model and preprocessing.}
Our primary classifier is ParticleNet-Lite~\cite{Qu2020ParticleNet}. Each jet retains at most the 100 constituents with largest original transverse momentum, and the model is trained on the standard ParticleNet input features derived from the constituent four-momenta. The degree-three polynomial classifier of Section~\ref{subsec:jet-tagging} is built from coordinate-dependent jet aggregates of the same constituent four-momenta.

\newpage

\section{Additional experiments of unit-rescaling group}
\label{appendix:unit_group_other_exp}

\paragraph{Numerical realization of the projection.}
\label{sec:units-quadrature}

Proposition 4.1 evaluates \(Q_v\) in closed form by deleting the coefficients of monomials in the wrong units sector.  Outside the Laurent-polynomial class that closed form is unavailable and the projection must instead be computed as an
average over the compact torus, so we first check that such an average
reproduces the algebraic operator.

We compare three evaluations of the same operator.  Exact coefficient filtering
applies \(Q_v\) by deleting wrong-sector coefficients.  Deterministic torus
quadrature averages over a finite roots-of-unity grid.  Monte Carlo projection
averages random phases \(z=e^{i\theta}\), where \(M\) denotes the total number
of model evaluations.  The comparison uses 20 frozen fitted predictors: five
seeds for each combination of force regression or Reynolds classification and
fixed-unit or unit-randomized training.  Writing \(f\) for a frozen fitted
predictor, the error of a numerical approximation \(\widehat Q\) is
\(\|\widehat Qf-Q_vf\|_2/\|Q_vf\|_2\) over the evaluation set.

\begin{table}[ht!]
\centering
\scriptsize
\setlength{\tabcolsep}{5pt}
\begin{tabular}{@{}lccc@{}}
\hline
Operator & Mean error & Maximum error \\
\hline
Deterministic & \(7.32\times10^{-16}\) 
& \(1.11\times10^{-15}\) \\
\(M=32\)  & \(0.0681\)  & \(0.449\) \\
\(M=128\) & \(0.0325\)  & \(0.203\) \\
\(M=256\) & \(0.0206\)  & \(0.109\) \\
\hline
\end{tabular}
\caption{Relative prediction error against exact coefficient filtering.
Monte Carlo rows aggregate combinations of task, training regime, seed, and
independent phase sequence.}
\label{tab:units-quadrature}
\end{table}

For every tested fit, deterministic quadrature reproduced coefficient filtering
to machine precision.  Monte Carlo error decreased as \(M\) increased but
remained heterogeneous across predictors: at \(M=256\), the mean was
\(0.0206\) and the maximum was \(0.109\).

Hence, averaging over the compact torus realizes the algebraic Weyl projection: the deterministic quadrature is exact up to floating-point error.  The Monte Carlo
realization converges slowly enough that its maximum error over predictors is
still above \(10^{-1}\) at 256 evaluations, which bounds what a sampled average
can be expected to deliver for predictors outside the Laurent class.  This
experiment concerns the numerical operator, not its predictive benefit.

\paragraph{Robustness under unit changes.}
\label{sec:units-robustness}

Section 4.5.2 reports a single training-set size and a single dictionary size.
Here we vary both, to test how the tradeoff between fixed-unit accuracy and
robustness under unit changes depends on them.  For drag-force regression, we
vary the number of labeled training examples \(N\in\{50,100,300,1000,3000\}\) (Table \ref{tab:units-robustness})
and the number of non-invariant features \(W\in\{0,25,100,250,500\}\) (Table \ref{tab:units-robustness-w-sweep}), that is,
dictionary features whose units weight \(Ua\) does not match the output weight
\(v_F\) and which \(Q_{v_F}\) therefore deletes.  We compare raw ridge, its
exact coefficient projection, the frame average, and a direct invariant
regression fitted on the invariant features alone.  We also test an empirical
\(L^2(\mu)\) approximation, where \(\mu\) is an input distribution: a ridge
regression on the invariant features is fitted to predictions of the frozen raw
model on a separate unlabeled calibration sample.  Unlike exact projection,
this refit may change the coefficients of the invariant features.

\begin{table}[ht]
\centering
\scriptsize
\setlength{\tabcolsep}{3pt}
\begin{tabular}{@{}lrlccc@{}}
\hline
Training & \(N\) & Model, \(W=500\)
& \shortstack{Fixed units}
& \shortstack{Transformed units\\(small shift)}
& \shortstack{Transformed units\\(large shift)} \\
\hline
Fixed units & \(50\) & Baseline
& \(1.03\pm0.594\)
& \((3.83\pm7.52)\times10^{11}\)
& \((1.08\pm2.21)\times10^{28}\) \\
& & Weyl-Projected
& \(0.776\pm0.0616\)
& \(0.770\pm0.0643\)
& \(0.776\pm0.0602\) \\
& & Empirical \(L^2(\mu)\)
& \(0.662\pm0.499\)
& \(0.629\pm0.453\)
& \(0.753\pm0.741\) \\
& & Invariant regression
& \(0.0224\pm0.0215\)
& \(0.0222\pm0.0204\)
& \(0.0229\pm0.0211\) \\
& & Frame average \(m=4\)
& \(11.7\pm9.38\)
& \(13.5\pm13.6\)
& \(12.8\pm10.5\) \\
\cline{2-6}
& \(3000\) & Baseline
& \(0.0167\pm0.00563\)
& \((5.01\pm14.0)\times10^{11}\)
& \((1.74\pm4.17)\times10^{27}\) \\
& & Weyl-Projected
& \(0.291\pm0.0747\)
& \(0.287\pm0.0755\)
& \(0.292\pm0.0751\) \\
& & Empirical \(L^2(\mu)\)
& \(0.00768\pm0.00244\)
& \(0.00786\pm0.00294\)
& \(0.00807\pm0.00288\) \\
& & Invariant regression
& \(0.00435\pm0.00185\)
& \(0.00422\pm0.00151\)
& \(0.00436\pm0.00166\) \\
& & Frame average \(m=4\)
& \(2.15\pm2.16\)
& \(1.67\pm1.06\)
& \(2.37\pm2.18\) \\
\hline
\shortstack{Transformed units}
& \(50\) & Baseline
& \(0.367\pm0.471\)
& \(104\pm229\)
& \((2.21\pm6.94)\times10^{17}\) \\
\shortstack{(small shift)} & & Weyl-Projected
& \(0.314\pm0.207\)
& \(0.309\pm0.202\)
& \(0.313\pm0.210\) \\
& & Empirical \(L^2(\mu)\)
& \(287\pm630\)
& \(287\pm620\)
& \(267\pm575\) \\
& & Invariant regression
& \(0.0224\pm0.0215\)
& \(0.0222\pm0.0204\)
& \(0.0229\pm0.0211\) \\
& & Frame average \(m=4\)
& \(0.794\pm0.794\)
& \(0.661\pm0.580\)
& \(0.726\pm0.695\) \\
\cline{2-6}
& \(3000\) & Baseline
& \(0.0197\pm0.0118\)
& \(0.0623\pm0.0726\)
& \((2.02\pm5.37)\times10^{12}\) \\
& & Weyl-Projected
& \(0.0241\pm0.0159\)
& \(0.0232\pm0.0157\)
& \(0.0234\pm0.0168\) \\
& & Empirical \(L^2(\mu)\)
& \(0.0502\pm0.0477\)
& \(0.0663\pm0.0920\)
& \(0.0600\pm0.0753\) \\
& & Invariant regression
& \(0.00435\pm0.00185\)
& \(0.00422\pm0.00151\)
& \(0.00436\pm0.00166\) \\
& & Frame average \(m=4\)
& \(0.0237\pm0.0115\)
& \(0.0235\pm0.0128\)
& \(0.0232\pm0.0127\) \\
\hline
\end{tabular}
\caption{Test NRMSE at \(W=500\) for two training-set sizes.  Rows give the
training regime and the model; columns give the unit system of the test data.
Entries are the mean \(\pm\) sample standard deviation over seeds.}
\label{tab:units-robustness}
\end{table}

\begin{table}[ht]
\centering
\scriptsize
\setlength{\tabcolsep}{3pt}
\begin{tabular}{@{}lrlccc@{}}
\hline
Training & \(W\) & Model, \(N=3000\)
& \shortstack{Fixed units}
& \shortstack{Transformed units\\(small shift)}
& \shortstack{Transformed units\\(large shift)} \\
\hline
Fixed units & \(25\) & Baseline
& \(0.00532\pm0.00281\)
& \((8.17\pm18.1)\times10^{5}\)
& \((3.24\pm6.23)\times10^{17}\) \\
& & Weyl-Projected
& \(0.00506\pm0.00157\)
& \(0.00488\pm0.00146\)
& \(0.00519\pm0.00170\) \\
& & Empirical \(L^2(\mu)\)
& \(0.00452\pm0.00228\)
& \(0.00425\pm0.00176\)
& \(0.00449\pm0.00197\) \\
& & Invariant regression
& \(0.00435\pm0.00185\)
& \(0.00422\pm0.00151\)
& \(0.00436\pm0.00166\) \\
& & Frame average \(m=4\)
& \(0.281\pm0.182\)
& \(0.266\pm0.210\)
& \(0.310\pm0.236\) \\
\cline{2-6}
& \(250\) & Baseline
& \(0.0154\pm0.00737\)
& \((1.74\pm3.71)\times10^{11}\)
& \((2.16\pm5.64)\times10^{26}\) \\
& & Weyl-Projected
& \(0.0663\pm0.0494\)
& \(0.0651\pm0.0485\)
& \(0.0656\pm0.0482\) \\
& & Empirical \(L^2(\mu)\)
& \(0.00808\pm0.00331\)
& \(0.00853\pm0.00377\)
& \(0.00880\pm0.00405\) \\
& & Invariant regression
& \(0.00435\pm0.00185\)
& \(0.00422\pm0.00151\)
& \(0.00436\pm0.00166\) \\
& & Frame average \(m=4\)
& \(1.58\pm1.05\)
& \(1.70\pm1.28\)
& \(1.91\pm1.68\) \\
\hline
\shortstack{Transformed units}
& \(25\) & Baseline
& \(0.00436\pm0.00182\)
& \(0.00459\pm0.00159\)
& \((9.76\pm21.1)\times10^{7}\) \\
\shortstack{(small shift)} & & Weyl-Projected
& \(0.00437\pm0.00182\)
& \(0.00421\pm0.00152\)
& \(0.00444\pm0.00165\) \\
& & Empirical \(L^2(\mu)\)
& \(0.00463\pm0.00184\)
& \(0.00474\pm0.00179\)
& \(0.00486\pm0.00187\) \\
& & Invariant regression
& \(0.00435\pm0.00185\)
& \(0.00422\pm0.00151\)
& \(0.00436\pm0.00166\) \\
& & Frame average \(m=4\)
& \(0.00440\pm0.00187\)
& \(0.00425\pm0.00156\)
& \(0.00446\pm0.00167\) \\
\cline{2-6}
& \(250\) & Baseline
& \(0.0138\pm0.0117\)
& \(0.0380\pm0.0250\)
& \((2.41\pm6.00)\times10^{12}\) \\
& & Weyl-Projected
& \(0.0154\pm0.0129\)
& \(0.0146\pm0.0119\)
& \(0.0145\pm0.0120\) \\
& & Empirical \(L^2(\mu)\)
& \(0.0402\pm0.0515\)
& \(0.0534\pm0.0963\)
& \(0.0482\pm0.0796\) \\
& & Invariant regression
& \(0.00435\pm0.00185\)
& \(0.00422\pm0.00151\)
& \(0.00436\pm0.00166\) \\
& & Frame average \(m=4\)
& \(0.0175\pm0.0115\)
& \(0.0167\pm0.0117\)
& \(0.0168\pm0.0111\) \\
\hline
\end{tabular}
\caption{Test NRMSE at \(N=3000\) for two non-invariant feature counts.
Rows give the training regime and the model; columns give the unit system
of the test data. Entries are the mean \(\pm\) sample standard deviation
over ten seeds.}
\label{tab:units-robustness-w-sweep}
\end{table}

\section{Weyl unitary trick with Lorentz group on vector valued function}
\label{apsubsec:weyl-lorentz}

In this section, we consider vector-valued outputs.  Let
\[
    F:(\mathbb C^d)^m\to\mathbb C^d.
\]
The Weyl unitarian projection equivariant Reynolds operator is
\begin{equation}
    (Q_K^{\mathrm{eq}}F)(U)
    :=
    \int_K k^{-1}F(kU)\,dk.
    \label{eq:lorentz-equivariant-average}
\end{equation}
This is the specialization of \eqref{eq:weyl-compact-average} with
\(\phi_C(k)U=kU\) and output representation \(\psi_C(k)=k\).  By the
same fixed-point argument used in the compact case,
\begin{equation}
    (Q_K^{\mathrm{eq}}F)(hU)
    =
    h(Q_K^{\mathrm{eq}}F)(U),
    \qquad h\in K.
    \label{eq:Weyl unitarian projection-equivariance}
\end{equation}

\begin{proposition}[Vector-valued Weyl unitarian projection average]
\label{prop:vector-valued-Weyl unitarian projection-average}
Let \(F:(\mathbb C^d)^m\to\mathbb C^d\) be polynomial, or more generally entire
holomorphic.  Then the Weyl unitarian projection average \(Q_K^{\mathrm{eq}}F\) is a
\(K\)-equivariant holomorphic map.  Moreover, it has the form
\begin{equation}
    (Q_K^{\mathrm{eq}}F)(U)
    =
    \sum_{j=1}^m \alpha_j(\Gamma(U))\,u_j,
    \label{eq:eq-output-span}
\end{equation}
where the coefficients \(\alpha_j\) are invariant scalar functions of the Gram data.
For polynomial \(F\), the \(\alpha_j\) may be chosen polynomial in the entries of
\(\Gamma(U)\); for entire \(F\), they may be chosen holomorphic on the relevant
Gram strata.
\end{proposition}

\begin{proof}
Equation~\eqref{eq:Weyl unitarian projection-equivariance} gives \(K\)-equivariance.  Since the
averaged map is holomorphic, \(K\)-equivariance extends to equivariance under the
complexified group \(O(d,\mathbb C)\).  The first fundamental theorem for orthogonal
covariants states that \(O(d,\mathbb C)\)-equivariant vector-valued polynomial maps
from \(m\) standard vectors to one standard vector are generated, as a module over
the invariant scalar functions, by the input vectors \(u_1,\ldots,u_m\).  Therefore
the output has the form \eqref{eq:eq-output-span}.  The holomorphic case follows by
applying the same argument to the homogeneous Taylor pieces and summing locally
uniformly.
\end{proof}

Thus the Weyl unitarian projection equivariant average always lies in the complex span of the
input vectors:
\[
    (Q_K^{\mathrm{eq}}F)(U)\in
    \operatorname{span}_{\mathbb C}\{u_1,\ldots,u_m\}.
\]
This is the vector-valued analogue of the Gram-matrix statement for scalar invariants.

On the locus where the Gram matrix is nondegenerate, the coefficients can be computed
directly.  Let \(U\) also denote the \(d\times m\) matrix whose \(j\)-th column is
\(u_j\), and set
\[
    G(U):=U^TU=\Gamma(U).
\]
Define
\begin{equation}
    b_i(U)
    :=
    \int_K F(kU)\cdot (ku_i)\,dk,
    \qquad i=1,\ldots,m.
    \label{eq:b-coefficients}
\end{equation}
Since \(k\) preserves the complex bilinear form,
\[
    b_i(U)
    =
    (Q_K^{\mathrm{eq}}F)(U)\cdot u_i.
\]
If \(G(U)\) is invertible and
\[
    (Q_K^{\mathrm{eq}}F)(U)=U\alpha(U),
\]
then
\[
    b(U)=G(U)\alpha(U),
\]
and hence
\begin{equation}
    \alpha(U)=G(U)^{-1}b(U),
    \qquad
    (Q_K^{\mathrm{eq}}F)(U)=U\,G(U)^{-1}b(U).
    \label{eq:eq-inverse-formula}
\end{equation}
More generally, a pseudoinverse formula is valid on strata where
\(\ker G(U)=\ker U\), equivalently where the bilinear form is nondegenerate on
\(\operatorname{span}_{\mathbb C}\{u_1,\ldots,u_m\}\).  On degenerate strata,
especially when null directions are present, the representation
\eqref{eq:eq-output-span} still holds, but the coefficients are not uniquely
determined by the pairings \(b_i\).  In such cases, a Moore--Penrose pseudoinverse
may be used as a numerical regularization, but it should not be interpreted as a
canonical algebraic formula.

Finally, suppose the original real output is Lorentz-vector-valued,
\[
    f:(\mathbb R^{1,d-1})^m\to\mathbb R^{1,d-1},
\]
and the complexified output is chosen so that
\[
    F(W(p_1),\ldots,W(p_m))=W(f(p_1,\ldots,p_m)).
\]
If the Weyl unitarian projection average restricts back to the Lorentz real slice, then
\eqref{eq:eq-output-span} becomes
\begin{equation}
    \widehat f(p_1,\ldots,p_m)
    =
    \sum_{j=1}^m
    \alpha_j\!\left(\big(\eta(p_i,p_j)\big)_{i,j=1}^m\right)
    p_j .
    \label{eq:lorentz-equivariant-output}
\end{equation}
Thus a Lorentz-equivariant vector output is built from the input four-vectors with
Lorentz-invariant scalar coefficients.  This is the equivariant analogue of
\eqref{eq:scalar-average-lorentz-gram}.

\end{document}